\documentclass[sn-mathphys,Numbered]{aux_bst/sn-jnl}% Math and Physical Sciences Reference Style

\usepackage{graphicx}%
\usepackage{multirow}%
\usepackage{amsmath,amssymb,amsfonts}%
\usepackage{amsthm}%
\usepackage{mathrsfs}%
\usepackage[title]{appendix}%
\usepackage{xcolor}%
\usepackage{textcomp}%
\usepackage{manyfoot}%
\usepackage{booktabs}%
\usepackage{algorithm}%
\usepackage{algorithmicx}%
\usepackage{algpseudocode}%
\usepackage{listings}%
\usepackage{placeins}
\usepackage {makecell} %//in table
\usepackage{multirow}%multirow in table
\usepackage{algpseudocode}
\usepackage{url}
\usepackage{bm}
\usepackage{color}
\usepackage{MnSymbol}
\usepackage{ulem}
\DeclareSymbolFont{ugmL}{OMX}{mdugm}{m}{n}
\DeclareMathAccent{\wideparen}{\mathord}{ugmL}{"F3} %arc length
\usepackage{cases} %mutiple equations
\usepackage{wrapfig}
\usepackage{todonotes}
\usepackage{soul}
\usepackage{pifont}
\usepackage{romannum}
\usepackage[final]{changes}
\definechangesauthor[name={Yuyan2}, color=orange]{2}
\theoremstyle{thmstyleone}%
\newtheorem{theorem}{Theorem}%  meant for continuous numbers

\newtheorem{lemma}{Lemma}

\newtheorem{assumption}{Assumption}

\theoremstyle{thmstyletwo}%

\theoremstyle{thmstylethree}%
\usepackage{placeins} % To control float placement with \FloatBarrier
\usepackage{geometry}
\usepackage{caption}  % For caption customization

\begin{document}

\title[Article Title]{\replaced{Pre-training with Fractional Denoising to Enhance Molecular Property Prediction}{Fractional Denoising: Enhancing Molecular Property Prediction with Chemical Priors}}
% first submission: Fractional Denoising: Enhancing Molecular Property Prediction with Chemical Priors
% Fractional Denoising: Enhancing Molecular Property Prediction through Chemical-Aware Conformational Distribution Modeling}
% Fractional Denoising for Molecular Property Prediction with Chemical Distributional Prior}


\author[2,3,1]{\fnm{Yuyan} \sur{Ni}}\email{niyuyan17@mails.ucas.ac.cn}
\equalcont{Equal contribution.}\equalconti{Work was done while Yuyan Ni was a research intern at AIR.} 
\author[1]{\fnm{Shikun} \sur{Feng}}\email{fsk21@mails.tsinghua.edu.cn}
\equalcont{Equal contribution.}
\author[1]{\fnm{Xin} \sur{Hong}}\email{hongxin@air.tsinghua.edu.cn}
\author[5,4,3]{\fnm{Yuancheng} \sur{Sun}}\email{sunyuancheng2021@ia.ac.cn}
\author[1]{\fnm{Wei-Ying} \sur{Ma}}\email{maweiying@air.tsinghua.edu.cn} 
\author[2,3]{\fnm{Zhi-Ming} \sur{Ma}}\email{mazm@amt.ac.cn}
\author[4]{\fnm{Qiwei} \sur{Ye}}\email{chivee.ye@gmail.com}
\author*[1,4]{\fnm{Yanyan} \sur{Lan}}\email{lanyanyan@air.tsinghua.edu.cn} 

\affil[1]{\orgdiv{Institute for AI Industry Research (AIR)}, \orgname{Tsinghua University}, \orgaddress{\city{Beijing}, \country{China}}}
\affil[2]{\orgdiv{Academy of Mathematics and Systems Science}, \orgname{Chinese Academy of Sciences}, \orgaddress{\city{Beijing}, \country{China}}}
\affil[3]{\orgname{University of Chinese Academy of Sciences}, \orgaddress{\city{Beijing}, \country{China}}}
\affil[4]{\orgname{Beijing Academy of Artificial Intelligence}, \orgaddress{\city{Beijing}, \country{China}}}
\affil[5]{\orgdiv{Institute of Automation},\orgname{Chinese Academy of Sciences}, \orgaddress{\city{Beijing}, \country{China}}}


\abstract{
Deep learning methods have been considered promising for accelerating molecular screening in drug discovery and material design. Due to the limited availability of labelled data, various self-supervised molecular pre-training methods have been presented. While many existing methods utilize common pre-training tasks in computer vision (CV) and natural language processing (NLP), they often overlook the fundamental physical principles governing molecules. In contrast, applying denoising in pre-training can be interpreted as an equivalent force learning, but the limited noise distribution introduces bias into the molecular distribution. To address this issue, we introduce a molecular pre-training framework called fractional denoising (Frad), which decouples noise design from the constraints imposed by force learning equivalence. In this way, the noise becomes customizable, allowing for incorporating chemical priors to significantly improve molecular distribution modeling. Experiments demonstrate that our framework consistently outperforms existing methods, establishing state-of-the-art results across force prediction, quantum chemical properties, and binding affinity tasks. The refined noise design enhances force accuracy and sampling coverage, which contribute to the creation of physically consistent molecular representations, ultimately leading to superior predictive performance.
}
% Experiments show that our framework is able to achieve new state-of-the-art results on the tasks of force, quantum chemical properties, and binding affinity prediction. The refined noise design is able to strengthen force accuracy and sampling coverage, resulting in physical consistent molecular representation and superior prediction performance.
%%================================%%
%% Sample for structured abstract %%
%%================================%%




\maketitle

\section{Introduction}\label{sec1}
 % , including predicting physical, chemical, and binding properties,
%  FUTURE WORK:MATERIAL
% 第一段展开，denoising不做靶子。
% new version: 压缩 mask CL,denoise提到前面，显得contribution更多

Molecular Property Prediction (MPP) is a critical task for various domains such as drug discovery and material design~\cite{butler2018machinenat,wong2023discoverynat,li2020aimaterialsdiscovery,deng2023molecularpropertyprediction,stokes2020deepcell}. Traditional approaches, including first-principle calculations and wet-lab experiments, require a huge cost~\cite{dowden2019clinicalsuccessrates,galson2021failure}, thus prohibiting high-throughput screening of the molecules with desirable properties. Therefore, deep learning methods have been considered a promising way to reduce the cost and substantially accelerate the screening process~\cite{pyzer2022materialsdiscovery,schneider2018drugdiscovery}. 

The main difficulty faced by deep learning MPP methods is the scarcity of labeled molecular data. To alleviate the difficulty, various self-supervised molecular pre-training methods have been proposed to exploit intrinsic information in unlabeled molecular data. Existing pre-training strategies are largely inspired by computer vision (CV) \cite{chen2020simclr,he2022MAE} and natural language processing (NLP) \cite{dai2015semi,devlin2018bert} techniques, such as contrastive learning and masking. Contrastive methods aim to maximize the representation agreement of augmentations derived from the same molecule and minimizing the representation agreement of different molecules \cite{wang2022MolCLR,3DGCL,fang2023knowledgeCL,stark20223dinfomax,liu2021pre,li2022geomgcl,zeng2022accurate}, while masking, on the other hand, leverages the model to recover the complete molecule from a masked molecular string~\cite{zhang2021mgbert,ross2022large}, graph~\cite{xia2023Molebert,rong2020self} or structure~\cite{fang2022geometry}.

Unfortunately, directly borrowing the prevalent pre-training tasks in CV and NLP can be unsuitable for molecules, as they neglect the underlying chemical characteristics and physical principles of molecules. For instance, graph-level augmentations for contrastive learning, such as edge perturbation, and subgraph extraction, can significantly alter molecular properties, leading to ineffective representations for property prediction. As for masking, recovering masked atom type can be trivial when 3D coordinates are known~\cite{zhou2023unimol}. Therefore, it is crucial to incorporate chemical priors and the laws of physics in AI for scientific discovery \cite{wang2023scientific}, to design suitable pre-training methods for molecules and strengthen generalization and robustness of the molecular representations.

% Although recent advances of denoising methods~\cite{SheheryarZaidi2022PretrainingVD,luo2022one,ShengchaoLiu2022MolecularGP,jiao2022energy,feng2023may} proposed a physically interpretable pre-training task that is equivalent to learning approximate atomic forces of given conformations. However, their applicability is constrained by a fixed noise type for the sake of maintaining interpretability. To be specific, typical denoising methods first perturb the equilibrium conformations by coordinate Gaussian noise (CGN) and then train the neural networks to predict the noise from the noisy conformation. The noise type has to be set as coordinate Gaussian noise (CGN) with isotropic noise variance to preserve the force learning interpretation. Nevertheless, the isotropic CGN noise results in a biased molecular distribution, where the molecules exhibit isotropic vibrations around equilibrium positions. This deviates from the true molecular distribution, where molecules can not only vibrate in a small scale but also rotate along rotatable single bonds in a relatively large scale, as depicted in Figure~\ref{fig:overview}a. Consequently, the biased molecular distribution modeling in previous methods results in inaccurate force targets and a restricted sampling range in proximity to equilibriums. 

\begin{figure*}[ht!]
    \vspace{-20pt}
    \centering
    \includegraphics[width=0.9 \textwidth]{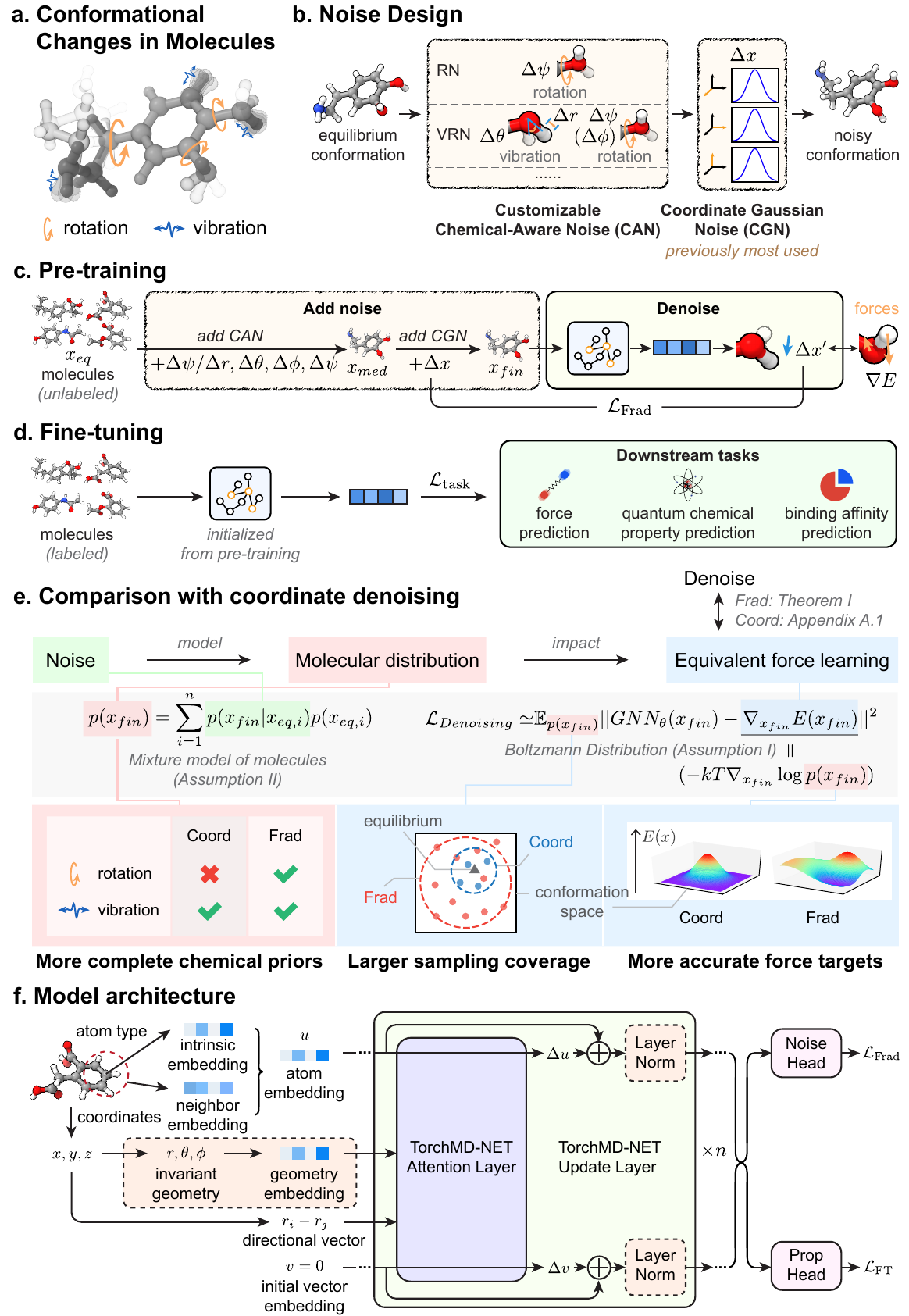}
    \vspace{-2pt}
    \caption{
    \textbf{Overview of Fractional denoising (Frad). a}. An illustration of the molecular conformational changes. The local structures can vibrate in small scale, while some single bonds can rotate flexibly. 
    \textbf{b}. The noise-adding process in the Frad framework. A two-phase hybrid random noise is applied to the equilibrium conformation, including the chemical-aware noise (CAN) that describes the molecular conformational changes and coordinate Gaussian noise (CGN). We present two versions of CAN.
    \textbf{c}. Pre-training process of Frad. The unlabeled molecular data is processed by adding noise and then utilized as the input of the graph neural networks to predict the CGN. This task is proved to be equivalent to learning the approximate atomic forces in the molecule. 
    % The better the noise design is aligned with true molecular conformation, the more accurate the forces are.
    \textbf{d}. Fine-tuning process of Frad. The GNN model inherits the pre-trained weights and continues to be updated together with a prediction head for specific downstream tasks.
    \textbf{e}. Advancements of Frad over with coordinate denoising methods (Coord)~\cite{SheheryarZaidi2022PretrainingVD,coordnoneq2023,ShengchaoLiu2022MolecularGP,jiao2022energy,luo2022one,feng2023may} through the perspective of chemical priors and physical interpretations. The noise of Frad is customizable, enabling capturing both rotations and vibrations in molecular conformation changes. Frad's superior modeling of molecular distribution further enabling larger sampling coverage and more accurate force targets in the equivalent force learning task, resulting in effective pre-training and improved downstream performance. 
   \added[id=2]{ \textbf{f}. An illustration of model architecture. The model primarily follows the TorchMD-NET framework, with our minor modifications highlighted in dotted orange boxes.}
    %(修改：不要画improved noisy node, 因为不同任务用了不同auxilary task)
    % \textbf{b}. The first phase of noise (Ph1 noise).  To learn a more realistic force field by Frad, we propose two example Ph1 noise distributions that better approximate the molecular distribution.
    % \textbf{c}. The whole Frad framework. GNNs are first pre-trained via Frad to learn representative features. Then in fine-tuning stage, the model inherit the pre-trained parameters of the GNN encoder as initialization and randomly initializes a prediction network for downstream molecular property predictions . Frad is also applied as an auxiliary task. The finetuning loss is a linear combination of downstream task loss and Frad loss.
     }\label{fig:overview}
     \vspace{-10pt}
    \end{figure*}

Recent advances of denoising methods~\cite{SheheryarZaidi2022PretrainingVD,luo2022one,ShengchaoLiu2022MolecularGP,jiao2022energy,feng2023may} have introduced a physically interpretable pre-training task, which is equivalent to learning approximate atomic forces of sampled noisy conformations. In these methods, equilibrium conformations are initially perturbed by noise, and neural networks are then trained to predict the noise based on the noisy conformation. The noise type in the previous denoising framework was restricted to set as coordinate Gaussian noise (CGN) with isotropic noise variance, to maintain the force learning interpretation. However, the use of isotropic CGN noise leads to a biased molecular distribution, focusing on isotropic vibrations around equilibrium positions, since molecules exhibit not only small-scale vibrations but also rotation along rotatable single bonds on a relatively large scale, as illustrated in Figure~\ref{fig:overview}a. Modeling this biased molecular distribution leads to inaccuracies in force targets and constraining the sampling range around equilibriums, as indicated by our theoretical analysis in Supplementary Information \ref{app:Chanllenges of coordinate denoising}, and ultimately hinders the model's performance on downstream tasks.

%chanllenge:物理解释同时与真实像。但真实分布unkown,解决idea是添加chemical priors 描述分布。基于idea提出XXX prior以CAN形式加入

Thus, the subsequent challenge lies in effectively modeling the comprehensive molecular distribution, while simultaneously preserving the essential physical interpretation of force learning.
Given the difficulty in modeling the true molecular distribution, we choose to characterize the distribution more comprehensively by introducing chemical priors about molecular distribution into noise design, which is prohibited in previous methods due to the restricted noise distribution.
Therefore, we propose a novel molecular pre-training framework called fractional denoising (Frad), which is proven to hold the force learning interpretation. Specifically, given an equilibrium molecular conformation, a hybrid noise of chemical-aware noise (CAN) and CGN is added and a noisy conformation is obtained, the model is trained to predict CGN from the noisy conformation. The term ``fractional'' refers to recovering a fraction of the entire noise introduced, with the necessity of the design discussed in Supplementary Information~\ref{sec:neccessity}. Notably, CAN is customizable enabling Frad to incorporate chemical priors to optimize molecular distribution modeling. 
% As a consequence, Frad successfully disentangles noise design from the requirements of force learning equivalence, thereby making room for incorporating chemical priors to optimize molecular distribution modeling. 
Inspired by the chemical priors that describe molecular conformational changes, we present two versions of CAN. Specifically, rotation noise (RN) is advocated to capture rotations of single bonds, while vibration and rotation noise (VRN) is put forward to reflect anisotropic vibrations.  
%1.考虑有两种prior  VRN RN缩短 2.fractional  最终tackle chanllenge既 又

% Thus Frad framework tackles the challenge of improving distribution modeling while maintaining interpretability, offering superior performance over existing denoising methods. 
To test how well Frad tackles the challenge, we carry out extensive experiments, validating superior performance over existing denoising methods in the following aspects.
Firstly, Frad allows for a more comprehensive exploration of the energy surface. Previous methods are vulnerable to generating irrational substructures like distorted aromatic rings and thus need to set the noise level quite low. Whereas Frad enables a broader exploration of the energy surface by a higher noise level on torsion angles of rotatable bonds. Empirical results show that Frad outperforms coordinate denoising (Coord) even with a perturbation scale 20 times larger.
Secondly, Frad learns more accurate atomic forces. As is proved in ``Result", the equivalent force learning target is derived from the modeled molecular distribution by Boltzman distribution. Thus by improving molecular distribution modeling, the force targets align better with the true atomic forces than previous methods, which is also validated by experiments.
Finally, the generality of the Frad framework permits introducing various chemical priors by alternating CAN, thereby accommodating a wide range of molecular systems and benefiting different downstream tasks. Therefore, Frad is able to achieve 18 new state-of-the-art on a total of 21 quantum chemical property and binding affinity prediction tasks from MD17, MD22, ISO17, QM9, and Atom3D datasets.

\section{Results}\label{sec:result}
\subsection{Frad framework}
To achieve physical-consistent self-supervised pre-training, we propose a novel fractional denoising framework that is equivalent to learning approximate atomic forces in molecules. \deleted{paradigm (depicted in Figure~\ref{fig:overview}) comprises two integral components: noise adding and denoising. }\added{The paradigm is depicted in Figure~\ref{fig:overview}.}

Given an equilibrium molecular conformation $x_{eq}$, a hybrid of chemical-aware noise (CAN) and coordinate Gaussian noise (CGN) are added, where the equilibrium conformation refers to the structure at local minima of the potential energy surface of the molecule. Then the model is trained to predict CGN from the noisy conformation, namely fractional denoising, as it recovers a portion of the introduced noise. Concretely, an 
equivariant graph neural network (GNN) is utilized to extract features from the noisy conformation, then a noise prediction head predicts the CGN from the features. 
 We employ TorchMD-NET~\cite{tholke2022torchmd} as the backbone model to obtain equivariant features from 3D molecular inputs\added[id=2]{, as shown in Figure~\ref{fig:overview}}. 
Notably, our theoretical analysis reveals that the task, irrespective of the distribution of CAN, possesses a force learning interpretation, whereas the CAN distribution affects the force targets and sampling distribution. Therefore, we meticulously design CAN to align with true molecular conformation distribution, resulting in more precise force targets and a wider sampling distribution when compared to existing denoising methods.  

% Specifically, the first phase of noise is a Chemical-aware Noise (CAN) that captures the unique characteristics of the molecular force field. The second phase of noise is a coordinate Gaussian noise (CGN) to guarantee the interpretation of atomic forces learning. This denoising task is not only applied as a pre-training task but is also utilized during the fine-tuning phase as an auxiliary task to enable positive transfer. 

During fine-tuning, we initialize the equivariant GNN from the pre-training weights and subsequently employ a distinct property prediction head tailored for each individual downstream task. The pre-trained GNN
weights along with the parameters in the prediction head continue to be updated under the supervision of downstream labels. More details about Frad are shown in “Methods”.
% Furthermore, we enhance an established fine-tuning technique known as Noisy Nodes, extensively utilized in quantum chemical property predictions but encountering convergence challenges in force prediction tasks. Specifically, we firstly decouple the input of the denoising and downstream tasks and then adopt Frad as an auxiliary task, concurrently updating the equivariant GNN alongside the primary downstream tasks. These modifications overcome previous convergence challenges, leading to improved results in force prediction tasks.
The entire pipeline is illustrated in Figure~\ref{fig:overview}.

\subsubsection{Atomic forces learning interpretation}
% This theoretical analysis provides valuable insight into why fractional denoising outperforms existing denoising methods \todo{demonstrate in next subsection} and demonstrates that fractional denoising is a versatile approach for learning intricate molecular force fields in a self-supervised way. 
We present a theorem establishing the equivalence between fractional denoising and the learning of atomic forces in molecules, thereby enhancing the interpretability of the denoising task. Unlike previous works, we seek to obtain the minimal conditions for this equivalence, affording greater flexibility in handling the noise distribution.
Prior to presenting the theorem, we first introduce relevant notations and assumptions that help to theoretically formulate the denoising task. 
% In statistical mechanics, under certain hypothesis, the conformation distribution of the molecule follows the Boltzmann distribution.
\begin{assumption}[Boltzmann Distribution~\cite{boltzmann1868studien}]\label{assump:Boltzmann_Dist} The probability that a conformation occurs relates to its energy in the following way:  
\begin{equation}
    % p_{physical}(x) \propto exp(-E_{physical}(x)),
    p(x) \propto exp(-\frac{E(x)}{kT}),
\end{equation}
where $x$ represents any conformation of the molecule, $E(x)$ denotes the potential energy function, $T$ signifies the temperature, and $k$ stands for the Boltzmann constant.
\end{assumption}
  We consider conformations in a fixed temperature, i.e. $kT$ is a constant. As an immediate consequence, a corollary arises: the score function of the conformation distribution equals the molecular forces up to a constant factor, i.e. $ \nabla_{x }\log p(x)= -\frac{1}{kT} \nabla_{x } E(x)$, where the score function is the gradient of the log of the probability density function of a probability distribution, and the gradient of the potential energy with respect to atomic coordinates is referred to as atomic forces or force field in our context.
   % $ (-kT\nabla_{x_{fin} }\log p(x_{fin}))=  \nabla_{x_{fin} } E(x)$
 
The pre-training dataset comprises equilibrium conformations of a large amount of molecules, where each conformation is represented by its atomic coordinates. For each molecule, we denote the equilibrium conformation as $x_{eq}$, the intermediate conformation after the introduction of CAN as $x_{med}$, and the final noisy conformation after adding the entire hybrid noise as $x_{fin}$. These variables \added{describe the Cartesian coordinates of the conformations and} are all real-valued vectors in $\mathbb{R}^{3N}$, where $N$ is the number of atoms constituting the molecule.

\begin{assumption}[Mixture Model of Molecules]\label{assump:mixture} 
The molecular distribution is approximated by a mixture model:
\begin{equation}\label{eq:assum mixture}
    p (x_{fin})=\sum_{i=1}^{n} p(x_{fin}|x_{eq,i})p(x_{eq,i}).
\end{equation} 
The component distribution $p (x_{fin}|x_{eq,i})$ is the distribution of hybrid noise, while the pre-determined mixing probability $p(x_{eq,i})$ characterizes the distribution of the equilibrium conformation and can be approximated by the sample distribution of the pre-training dataset. $n$ denotes the number of equilibrium conformations.
\end{assumption}
Therefore, we can sample a molecule by adding hybrid noise to the equilibrium conformations from the dataset. Also, in order to precisely model the molecular distribution, we should design the hybrid noise to better capture the true distribution landscape around the equilibriums, which is discussed in the subsequent section. 

Then we arrive at the main theorem that establishes the equivalence between fractional denoising and forces learning. 
\begin{theorem}[Learning Forces via Fractional Denoising]\label{thm:frad}
If the distribution of hybrid noise satisfies $p(x_{fin}|x_{med})\sim \mathcal{N}(x_{med},\tau^2  I_{3N})$ is a coordinate Gaussian noise (CGN), then fractional denoising is equivalent to learning the atomic forces that correspond to the approximate molecular distribution by Boltzmann Distribution. 
 % \vspace{-5pt}
     \begin{small} 
    \begin{subequations}\label{eq: thm}
     % \vspace{-3pt}
    \begin{equation}
        \mathbb{E}_{p (x_{fin}|x_{med})p(x_{med}|x_{eq})p(x_{eq})}||GNN_{\theta} (x_{fin}) - (x_{fin}-x_{med})||^2 \label{eq:frad target in thm}
    \end{equation}
     \begin{equation}
     \begin{aligned}
          \simeq  \mathbb{E}_{p (x_{fin})}||GNN_{\theta} (x_{fin}) - \nabla _{x_{fin}} E(x_{fin})||^2, \label{eq:learn ff target in thm}
     \end{aligned}
    \end{equation}      
    % \begin{equation}
    %  \begin{aligned}
    %       \mathcal{L}_{Denoising} \simeq & E_{p (x_{fin})}||GNN_{\theta} (x_{fin}) - \nabla _{x_{fin}} E(x_{fin})||^2\\
    %       &E_{p (x_{fin})}||GNN_{\theta} (x_{fin}) - (-\nabla _{x_{fin}} \log p(x_{fin}))||^2, 
    %  \end{aligned}
    % \end{equation}      
     % \vspace{-2pt}
    \end{subequations}
    \end{small} 
$\simeq$ denotes the equivalence as optimization objectives. \eqref{eq:frad target in thm} is the optimization objective of Frad. \eqref{eq:learn ff target in thm} is the regression task of fitting an approximate force field. 
\end{theorem}
The proof is provided in Supplementary Information~\ref{section:appendix proof thm1}. Since the force is a physically well-defined observable and a fundamental quantity for many downstream properties~\cite{chmiela2017machine,feng2023may}, Frad is capable of capturing intrinsic molecular laws and generalizing to various downstream tasks. More evidence verifying the effectiveness of force learning for molecular pre-training is provided in Supplementary Material.
 % \vspace{-6pt} 
% (TODO REF, add proof here and highlight the function of 2 noises, one for approximation, coord one for equivalence.).

\subsubsection{Chemical-aware noise design}\label{section:Hybrid noise design}
To faithfully model the true molecular distribution in equation~\eqref{eq:assum mixture}, ensuring realistic conformation sampling and precise force targets in equation~\eqref{eq:learn ff target in thm}, we should meticulously design the hybrid noise to capture the true distribution landscape surrounding the equilibriums. Fortunately, theorem~\ref{thm:frad} imposes constraints solely on the coordinate Gaussian noise (CGN) to establish the equivalence, thereby affording flexibility to Chemical-Aware Noise (CAN) in describing the molecular distribution.

One basic requirement for the hybrid noise is that $x_{eq}$ is the local maximum value of $p (x_{fin})$, so that the equilibrium conformations are local minima of the potential energy surface. A simple approach to satisfy this requirement is to employ CGN only, with no CAN introduced: $p(x_{med}|x_{eq})=I_{x_{med}=x_{eq}}$, $I$ is an indicator function, and $p(x_{fin}|x_{med})\sim \mathcal{N}(x_{med},\tau^2  I_{3N})$. By choosing a relatively small value for $\tau$, the requirement can be satisfied. This is exactly the prevalent coordinate denoising method~\cite{SheheryarZaidi2022PretrainingVD,luo2022one,feng2023may}.
Nevertheless, coordinate Gaussian noise is suboptimal for approximating the molecular distribution, as it can solely capture small-scale vibrations and does not account for rotations, which is significant for sampling broader low-energy conformations in the actual molecular distributions.
% , as it has an isotropic covariance. In reality, molecules can be composed of rigid and flexible parts, which should have varing noise variances. Specifically, rigid parts of a molecule, such as rings, double and triple bonds, should have smaller noise variance than flexible parts, such as rotatable single bonds. This is because small coordinate perturbations on the rigid parts can lead to high energy conformations, while altering the torsion angles of rotatable bonds (T-rot) does not cause significant energy changes.
% To further describe the anisotropic vibrations between atoms, we propose vibration and rotation noise (VRN) that perturbs bond lengths, bond angles, torsion angles, including that of rotatable bonds by Gaussian noise.

To address this limitation, we incorporate CAN to capture the intricate characteristics of molecular distributions. Initially, we introduce rotation noise (RN) that perturbs torsion angles of rotatable bonds by Gaussian noise, with the probability distribution given by  $p(\psi_{med}|\psi_{eq})\sim \mathcal{N}(\psi_{eq},\sigma^2  I_{m})$, where $\psi_{eq}, \psi_{med} \in [0,2\pi)^m$ are the torsion angles of rotatable bonds of $x_{eq}$, $x_{med}$ respectively. $m$ is the number of rotatable bonds in the molecule. The torsion angle refers to the angle formed by the intersection of two half-planes through two sets of three atoms, where two of the atoms are shared between the two sets. This model is termed Frad(RN). 
% The Frad framework can capture this complex characteristics of the molecular distribution. To achieve anisotropic noise covariance for rigid and flexible parts, we set the Ph1 noise to be Gaussian noise on T-rot, with the probability distribution given by  $p(\psi_{med}|\psi_{eq})\sim \mathcal{N}(\psi_{eq},\sigma^2  I_{m})$, where $\psi_{eq}, \psi_{med} \in [0,2\pi)^m$ are the T-rot of $x_{eq}$, $x_{med}$ respectively. $m$ is the number of T-rot in the molecule. This model is termed Frad(T). 

To provide a more comprehensive description of anisotropic vibrations, we model the conformation changes by vibration on bond lengths, bond angles, and torsion angles. As a result, we propose vibration and rotation noise (VRN) that perturbs bond lengths, bond angles, torsion angles, including that of rotatable bonds by independent Gaussian noise. The respective probability distributions are specified as follows: 
% The variances of each bond, angle and torsion angle are different.
   $ p(r_{med}|r_{eq})\sim \mathcal{N}(r_{eq}, \sigma_r^2I_{m_1}), p(\theta_{med}|\theta_{eq})\sim \mathcal{N}(\theta_{eq}, \sigma_{\theta}^2I_{m_2}), p(\phi_{med}|\phi_{eq})\sim \mathcal{N}(\phi_{eq}, \sigma_{\phi}^2I_{m_3})$, $p(\psi_{med}|\psi_{eq})\sim \mathcal{N}(\psi_{eq}, \sigma_{\psi}^2I_{m})$, where $r$, $\theta$, $\phi$, $\psi$ denote bond lengths, angles, torsion angles of non-rotatable bonds and rotatable bonds respectively, $m_1$, $m_2$, $m_3$, $m$ are their numbers in the molecule.

\subsection{Frad boosts the performances of property prediction}
 % Frad is pretrained on a large-scale molecular dataset PCQM4Mv2 \cite{nakata2017pubchemqc}, comprising 3.4 million organic molecules, each with one equilibrium conformation. 
 
To assess the efficacy of Frad in predicting molecular properties, we conducted a series of challenging downstream tasks. These tasks encompass atom-level force prediction, molecule-level quantum chemical properties prediction, and protein-ligand complex-level binding affinity prediction. Our model is systematically compared against established baselines, including pre-training approaches, as well as property prediction models without pre-training. In experimental results, we use the abbreviation `Coord' to refer to the coordinate denoising pre-training method in~\citet{SheheryarZaidi2022PretrainingVD}, which shares the same backbone as our model. \added[id=2]{The data splitting methods adhere to standard practices in the literature, where MD17, MD22, and QM9 adopt uniformly random splittings while  ISO17 and LBA utilize out-of-distribution splitting settings.} More information regarding datasets and baseline models is provided in ``Method". Detailed algorithms, metrics, hyperparameters, and ablation studies are provided in Supplementary Information~\ref{app: setting} and \ref{sec:supple exp}.

% In order to evaluate the effectiveness of Frad for molecular property prediction, we conduct several challenging downstream tasks, including atom-level force prediction, molecule-level quantum chemical properties prediction, and protein-ligand level binding affinity prediction. Detail descriptions of molecular datasets and baselines can be found in Supplementary material. 
% Our model is compared with existing baselines of pre-training approaches and property prediction models without pre-training. 

\subsubsection{Atom-wise force prediction}
% atom level +BAT Frad,+ MD dataset\
    \begin{figure}[t]
    \begin{center}
    \includegraphics[width=\textwidth]{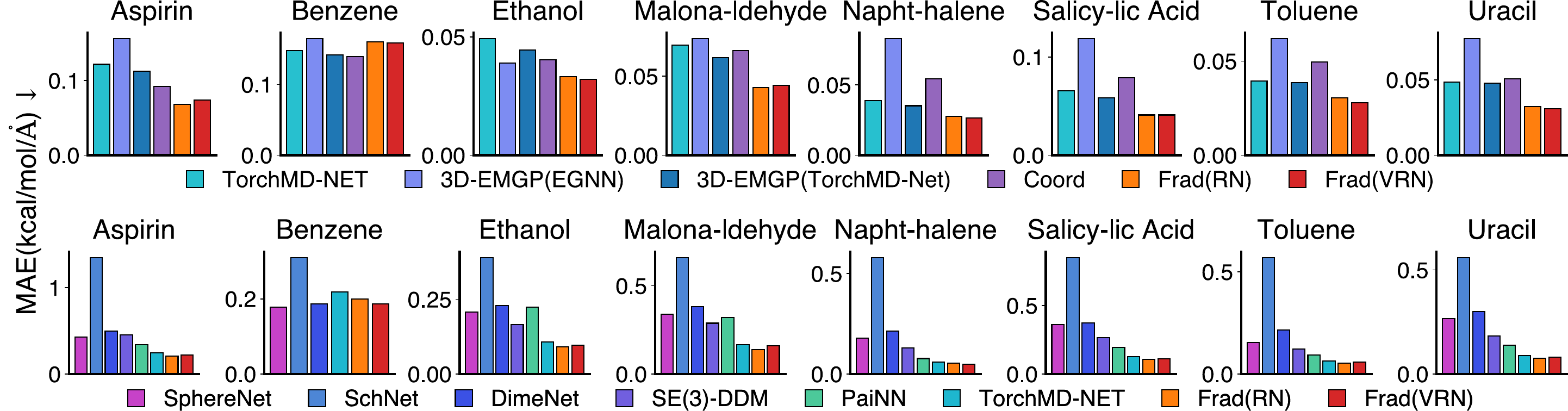}
    \caption{\textbf{Test performance of different models on force prediction tasks in the MD17 dataset.} To fairly compare with the existing methods, we assess Frad under two split settings commonly found in the literature. The top row are results using a large training data split and the bottom row are results with limited training data.  }
    % The bar chart displays the average performance (MAE, lower is better) of coordinate denoising and Frad methods on seven energy prediction tasks in QM9, including $\epsilon_{HOMO}$, $\epsilon_{LUMO}$, $\Delta\epsilon$, $U_0$, $U$, $H$, and $G$. To ensure fair representation of sample coverage, we also present the perturbation scale, defined as the mean absolute coordinate changes of all atoms after applying noise, in a line chart. The results show that Frad outperforms coordinate denoising, particularly when the perturbation scale is large, suggesting that Frad can effectively sample a broader range of conformations. .\footnotemark[1]}
     \label{fig:md17&analysis}
     \end{center}
    \end{figure}

   \begin{table*}[h]
\setlength{\tabcolsep}{0.35em}
    \caption{Test performance of different models on force prediction tasks on 7 molecules in the MD22 dataset as well as energy and force prediction tasks in the ISO17 dataset. The result is measured by root mean square error (RMSE) and mean absolute error (MAE) for each molecule in kcal/mol/$\mathring{\rm{A}}$ for force and in kcal/mol for energy. The best results are in bold.
    }
    \label{table:iso17 md22}
    \begin{center}
    \begin{footnotesize}
    \scalebox{1}{
    \begin{tabular}{llrrrrrrrr}
    \toprule
      \multirow{2}{*}{Datasets}&\multirow{2}{*}{Tasks}& \multicolumn{2}{c}{\makecell[c]{TorchMD-NET}}	 & 	\multicolumn{2}{c}{Coord(NonEq)} & \multicolumn{2}{c}{\makecell[c]{Frad(RN)}}	 & 	\multicolumn{2}{c}{Frad(VRN) } \\    
      \cmidrule(lr){3-4}\cmidrule(lr){5-6}\cmidrule(lr){7-8}\cmidrule(lr){9-10}
       & & RMSE $\downarrow$ & MAE $\downarrow$ & RMSE $\downarrow$& MAE $\downarrow$ & RMSE $\downarrow$ & MAE $\downarrow$ & RMSE $\downarrow$ & MAE $\downarrow$ \\
          \midrule      			        		               		             	
        \multirow{7}{*}{MD22}&Ac-Ala3-NHMe&0.154&0.113&0.139&0.102&0.101&0.076& \textbf{0.097}&\textbf{0.073}\\
        &DHA&0.200&0.144&0.185&0.135&0.112&0.083&\textbf{0.105}&\textbf{0.078}\\
        &AT-AT&0.394&0.274&0.392&0.288&0.372&0.266&\textbf{0.325}&\textbf{0.233} \\
        &Staychose&1.031&0.716&0.956&0.673&\textbf{0.313}&\textbf{0.201}  &0.356&0.231\\
        &AT-AT-CG-CG&1.141&0.758&0.994&0.657&0.705&0.364&\textbf{0.648}&\textbf{0.308} \\
        &Buckyball&3.207&1.820&3.087&1.751&1.456&0.528&\textbf{1.158}&\textbf{0.430}        \\
        &double-walled&5.004&3.236&3.944&2.515&2.429&1.510&\textbf{1.170}&\textbf{0.729}\\
        \midrule  
        \multirow{2}{*}{ISO17}&Energy&4.06&2.61&3.34&2.12&2.05&1.96&\textbf{1.38}&\textbf{1.29}\\
         &Force&2.64&1.45&2.01&1.21&1.88&1.20&\textbf{1.12}&\textbf{0.70}\\
    \bottomrule
    \end{tabular}}
    \end{footnotesize}
    \end{center}
\end{table*}

As fractional denoising theoretically engages in learning approximate forces, it is expected to confer advantages to downstream force learning tasks.
Therefore, we evaluate Frad's performance in predicting forces across various molecular dynamics datasets,
% namely MD17 [30], ISO17 [33], and MD22 [34].
% we evaluate the force prediction performance of Frad on several molecular dynamics datasets:
i.e.~MD17~\cite{chmiela2017machine}, ISO17~\cite{schutt2017schnetiso17}, and MD22~\cite{Chmiela2023MD22}. 
To assess Frad's capacity to extrapolate to a broader distribution of molecules, we adopt MD17 and MD22 datasets, which include molecules with respectively fewer and more atoms than the pre-training dataset on average. To further test the versatility of Frad on diversified chemical structures, we choose to use ISO17 dataset, which contains 129 isomers of C$_7$O$_2$H$_{10}$.

% SphereNet, SchNet, DimeNet and SE(3)-DDM split 1000 samples into training data and others uses 950.

 As numerous prior studies have adopted MD17 as their evaluation dataset, we compare Frad with several competitive pre-training methods and supervised models. Previous works diverge on data splitting, so we present the results in both large and limited training data scenarios in Figure \ref{fig:md17&analysis}. \added[id=2]{We do not provide results for PaiNN and SE(3)-DDM on Benzene because PaiNN does not report the result of Benzene and SE(3)-DDM reports the result of Benzene by using dataset~\cite{chmiela2018towards}, which is distinct from ours~\cite{chmiela2017machine}.} In both data splitting scenarios, Frad outperforms the baselines on 7 out of 8 molecules. Especially when compared to the denoising pre-training method with the same backbone as ours, i.e. 3D-EMGP~\cite{jiao2022energy} and Coord~\cite{SheheryarZaidi2022PretrainingVD}, our superiority is evident. This highlights the need for introducing chemical priors to accurately model molecular distribution in denoising. When the size of the training set is limited, Frad pre-training consistently enhances the backbone model TorchMD-NET by a large margin, indicating that Frad effectively learns the intrinsic principles of molecules and possesses excellent generalization ability. Regarding Benzene in the large training set setting, we notice overfitting during the fine-tuning of Frad, which is not observed in other molecules. This may be attributed to the relatively fixed structure of benzene, resulting in low-dimensional features that are susceptible to overfitting.

% MD17\cite{chmiela2017machine,chmiela2018towards,schutt2017quantum}
The results on ISO17 and MD22 are shown in Table~\ref{table:iso17 md22}. The performance of the backbone model TorchMD-NET is consistently improved by Frad pre-training. Coord(NonEq)~\citet{coordnoneq2023} refers to coordinate denoising pre-trained on nonequilibrium dataset ANI-1~\cite{smith2017ani} and ANI-1x~\cite{smith2020ani1X} with TorchMD-NET as the backbone. Although Coord(NonEq) is a competitive counterpart as nonequilibrium conformations are focused on in the force prediction tasks and Frad is pre-trained on equilibrium conformations, Frad still surpasses Coord(NonEq). This largely owes to our noise design that is able to augment conformations much further from the equilibrium structure than coordinate denoising by turning the torsion angles of the rotatable bonds. Moreover, the molecular distribution modeling of Frad is more precise than coordinate denoising centered on nonequilibrium conformations, leading to better force targets. An intriguing observation is that VRN outperforms RN in force prediction tasks, highlighting the necessity to model anisotropic vibrations. Finally, Frad exhibits consistent advantages across the two datasets, indicating its broad applicability to complex structures and larger molecules.

\subsubsection{Molecule-wise quantum chemical properties prediction}
% - (molecule level) +BAT Frad,+QM dataset
  \begin{table*}[h]
% \vspace{-7pt}
\centering
\footnotesize
    \caption{Test performance of different models on 12 tasks in QM9 dataset. The upper half of the models are supervised learning methods and the lower half are pre-training methods. The mean absolute error (MAE $\downarrow$) is reported. The best and second best results are in bold and underlined, respectively. }
    \label{table:qm9}
    % \begin{center}
    % \begin{footnotesize}
    % \begin{sc}
    % \scalebox{0.9}{
\setlength{\tabcolsep}{0.2em}
    \begin{tabular}{lrrrrrrrrrrrr}
    \toprule
    	Models & \makecell[c]{$\mu$ \\(D)}	& 	\makecell[c]{$\alpha$\\ ($a_0^3$)}		&  \makecell[c]{HOMO \\(meV)}		& \makecell[c]{LUMO\\ (meV)}		& \makecell[c]{Gap\\ (meV)}	& \makecell[c]{$\langle R^2\rangle$ \\($a_0^2$)}	& \makecell[c]{ZPVE\\ (meV)	}	& \makecell[c]{$U_0$ \\ (meV)}		& \makecell[c]{$U$ \\ (meV)}		& \makecell[c]{$H$ \\ (meV)}		& \makecell[c]{$G$\\ (meV)} & \makecell[c]{$C_v$\\ ($\frac{cal}{mol K}$)	}
     \\
    \midrule
    SchNet & 	0.033 & 	0.235	 & 41.0 & 	34.0 & 63.0	 & \underline{0.070}	 & 1.70	 & 14.00	 & 19.00	 & 14.00	 & 14.00	 & 0.033\\
    E\deleted{(n-)}GNN & 	0.029	 & 0.071	 & 29.0	 & 25.0	 & 48.0	 & 0.110	 &   1.55	 & 11.00	 & 12.00	 & 12.00	 & 12.00	 & 0.031\\
    DimeNet++	 & 0.030	 & 0.043	 & 24.6	 & 19.5	 & 32.6	 & 0.330	 & 1.21	 & 6.32	 & 6.28 & 	6.53	 & 7.56	 & 0.023\\
    PaiNN	 & 0.012	 & 0.045	 & 27.6	 & 20.4	 & 45.7	 & \underline{0.070}	 & 1.28	 & 5.85	 & 5.83	 & \underline{5.98}	 & 7.35	 & 0.024\\
    SphereNet & 0.027 & 0.047 & 23.6 & 18.9 & 32.3  & 0.290 & \textbf{1.12} & 6.26&  7.33 & 6.40 &8.00 &0.022\\ 
    TorchMD-NET & 0.011 & 0.059 & 20.3 & 18.6 & 36.1 & \textbf{0.033} & 1.84 & 6.15 & 6.38 & 6.16 & 7.62 & 0.026 \\
   \midrule % \hline
    % ChemRL-GEM  & & &\multicolumn{3}{|c|}{199.726538		}  & &  & &  & & \\
    % Uni-mol  & & &\multicolumn{3}{|c|}{	127.073969	}  & &  & &  & & \\
    % \hline
    Transformer-M &	0.037 &		\underline{0.041} &		\underline{17.5} &		16.2 &		\textbf{27.4} &		0.075 &		\underline{1.18} &		9.37 &		9.41 &		9.39 &		9.63 &		0.022
     \\
     SE(3)-DDM 	 &	0.015	 &	0.046 &		23.5	 &	19.5	 &	40.2	 &	0.122	 &	1.31	 &	6.92 &		6.99	 &	7.09	 &	7.65	 &	0.024
     \\
    3D-EMGP &	0.020	 &	0.057	 &	21.3	 &	18.2	 &	37.1	 &	0.092	 &	1.38	 &	8.60 &		8.60	 &	8.70	 &	9.30	 &	0.026
    \\
    % DP-GNS-TAT*	 &	0.016	 &	0.040	 &	14.9	 &	14.7	 &	22.0	 &	0.44	 &	1.018	 &	5.76	 &	5.76	 &	5.79	 &	6.90	 &	0.020
    % \\
    Coord	 &	0.012	 &	0.052	 &	17.7	 &	14.3	 &	31.8	 &	0.450	 &	1.71	 & 6.57  &		 6.11  &		 6.45  &		 6.91 
    	 &	\textbf{0.020}
    \\  
     \makecell[l]{Frad(VRN)} &		\underline{0.011} &		0.042	 &	17.9	 &	\underline{13.8}	 &	\underline{27.7}	 &	0.354 &		1.63	 &	\underline{5.41} &	\textbf{5.35}	& 6.01 &	\textbf{6.03}
    	 &	0.021 \\
    \makecell[l]{Frad(RN)} &		\textbf{0.010} &		\textbf{0.037}	 &	\textbf{15.3}	 &	\textbf{13.7}	 &	27.8	 &	0.342 &		1.42	 &	\textbf{5.33} &	\underline{5.62}	& \textbf{5.55} &	\underline{6.19}
    	 &	\textbf{0.020} \\    
    \bottomrule
    \end{tabular}
    % }
    % \end{footnotesize}
    % \end{center}
    % \vskip -0.2in
\end{table*}
To verify whether Frad can consistently achieve competitive results on different properties, we evaluate Frad(RN) and Frad(VRN) on 12 tasks in QM9 dataset~\cite{ruddigkeit2012enumeration,ramakrishnan2014quantum}. The results of Frad together with pre-training and supervised baselines are summarized in Table \ref{table:qm9}. In general, we exceed the supervised and pre-training methods and achieve a new state-of-the-art for 9 out of 12 targets. In addition, we make remarkable improvements on the basis of the backbone model TorchMD-NET on 11 targets, indicating the knowledge learned by Frad pre-training is helpful for most downstream tasks. Moreover, we have an evident advantage over the denoising pre-training methods in the lower half of the table. Especially, our Frad achieves or surpasses the results of the coordinate denoising approach with the same backbone TorchMD-NET in all 12 tasks, revealing that the introduced distributional chemical prior contributes to multiple categories of properties. We also note that VRN and VN produce comparable performance. We speculate the modeling of anisotropic vibration noise may not be crucial for tasks within the QM9 dataset that are less sensitive to input conformation compared to force prediction tasks.
% Here DP-TorchMD-NET is trained with hyperparameters in the code of \citet{SheheryarZaidi2022PretrainingVD}. A comparison with strictly aligned setting between coordinate denoising and Frad is in section \ref{section:ablation1}.
% Note that QM9 contains multiple categories of equilibrium properties, including thermodynamic properties, spatial distribution of electrons and states of the electrons. We speculate an accurate atomic forces learning can not only assist energy prediction but may enhance the atomic charge prediction and its related properties as well.  

\subsubsection{Complex-wise binding affinity prediction}
%without using noisy nodes
 Protein-ligand binding affinity prediction aims to predict the interaction strength between proteins and ligands. We assess the performance of Frad on ligand binding affinity (LBA) task in Atom3D dataset~\cite{atom3d}, where the conformation of the protein-ligand complexes and binding affinity labels are provided. Following Atom3D preprocessing~\cite{atom3d}, Frad extracts the binding pocket from the protein and utilizes pocket-ligand complex structure as input to predict the binding affinity value. The results are shown in Table~\ref{table:lba}. Frad beats the sequence-based and structure-based baselines tailored for protein representation, demonstrating that Frad is a universal representation learning method for both small molecules and proteins. The pre-training baseline SE(3)-DDM~\cite{ShengchaoLiu2022MolecularGP}, \added[id=2]{which does not report the result in split by identity 60\%, }is an equivariant coordinate denoising method that denoises the Gaussian noise on pairwise atomic distance. Likewise, Frad surpasses the traditional denoising method, again indicating the significance of introduced chemical priors.

  \begin{table}[ht]
\centering
\footnotesize
    \caption{Performance comparison of various methods on LBA dataset under two different split settings with $30\%$ and $60\%$ protein sequence identity between training and test sets. \added[id=2]{The best results are in bold.}}
    \label{table:lba}
    % \begin{footnotesize}
\setlength{\tabcolsep}{0.6em}
    \begin{tabular}{llrrrrrr}
        \toprule
        & & \multicolumn{3}{c}{LBA 30\%} & \multicolumn{3}{c}{LBA 60\%} \\
        \cmidrule(lr){3-5} \cmidrule(lr){6-8}
        Methods & Model & RMSE$\downarrow$ & Pearson$\uparrow$ & Spearman$\uparrow$ & RMSE$\downarrow$ & Pearson$\uparrow$ & Spearman$\uparrow$ \\
        \midrule
        \multirow{3}{*}{\shortstack{Sequence \\ based DL}} & DeepDTA & 1.866 & 0.472 & 0.471 & 1.762 & 0.666 & 0.663 \\
        & TAPE & 1.890 & 0.338 & 0.286 & 1.633 & 0.568 & 0.571 \\
        & ProtTrans & 1.544 & 0.438 & 0.434 & 1.641 & 0.595 & 0.588 \\
        \midrule
        \multirow{5}{*}{\shortstack{Structure \\ based DL}} & Atom3D-CNN & 1.416 & 0.550 & 0.553 & 1.621 & 0.608 & 0.615 \\
        & Atom3D-ENN & 1.568 & 0.389 & 0.408 & 1.620 & 0.623 & 0.633 \\
        & Atom3D-GNN & 1.601 & 0.545 & 0.533 & 1.408 & 0.743 & 0.743 \\
        & Holoprot & 1.464 & 0.509 & 0.500 & 1.365 & 0.749 & 0.742 \\
        & ProNet & 1.463 & 0.551 & 0.551 & 1.343 & 0.765 & 0.761 \\
        \midrule
        \multirow{2}{*}{\shortstack{Pre-training \\ Methods}} 
        % & DeepAffinity & 1.893 & 0.415 & 0.426 & - & - & - \\
        % & SMT-DTA & 1.574 & 0.458 & 0.447 & 1.347 & 0.758 & 0.754 \\
        & SE(3)-DDM & 1.451 & 0.577 & 0.572 & - & -  & -  \\
        & Frad & \textbf{1.365} & \textbf{0.59\replaced{9}{88}} & \textbf{0.577\deleted{2}} & \textbf{1.213} & \textbf{0.804\deleted{4}} & \textbf{0.801\deleted{4}} \\
        \bottomrule
    \end{tabular}
    % \end{footnotesize}
\end{table}

%Setting splitting dataset problem definition  references...
    %force

% We generate conformers using RDKit~\cite{landrum2013rdkit} as pre-training data, which is less accurate but much faster than DFT.We then pre-train Frad($\tau=0.04, \sigma=2$) and Coord ($\tau=0.04$) on this dataset. 

% \subsection{Ablation Study}
\subsection{Frad is robust to inaccurate conformations}
A large pre-training dataset with equilibrium conformations constitutes a fundamental prerequisite for effective 3D molecular pre-training. However, constructing such a dataset can be costly, as it typically requires the use of density functional theory (DFT) to calculate the equilibrium conformations. As a result, we turn to explore the sensitivity of the model to the accuracy of the pre-train data, evaluating whether Frad maintains efficacy when conformations are computed using fast yet less accurate methods. We employ RDKit~\cite{landrum2013rdkit} for regenerating 3D conformers on the original PCQM4Mv2 pre-training dataset, which is less accurate but much faster than DFT. Subsequently, we perform pre-training on Frad($\tau=0.04, \sigma=2$) and Coord ($\tau=0.04$) using this inaccurate dataset. Shown in \ref{fig:md17&analysis2} are their downstream performance in HOMO, LUMO, Gap prediction tasks on the QM9 dataset compared to the models trained on the original pre-train dataset. 

The results show that pre-training on inaccurate conformations leads to larger mean absolute errors. However, denoising pre-training methods remain effective and outperform the model trained from scratch. In particular, Frad consistently outperforms Coord. Intriguingly, Frad trained with inaccurate conformations even surpasses Coord trained with accurate conformations. These findings verify that Frad is a highly effective pre-training model, even when using inaccurate conformations, making it possible to pre-train on a larger scale while using less accurate pre-training datasets.

\subsection{Comparisons to coordinate denoising methods}
As discussed in ``Introduction", coordinate denoising methods face the challenge of biased molecular distribution modeling, leading to restricted sampling coverage and inaccurate force targets. In this section, we quantitatively validate that Frad can augment the sampling coverage as well as enhance force accuracy, thereby yielding superior downstream performance.
\subsubsection{Frad achieves higher force accuracy} \label{sec:force accuracy}
% +BAT Frad
% In section \ref{section:Hybrid noise design}, we design hybrid noise to better model the true molecular distribution as compared to prior coordinate denoising methods. Here we quantitatively investigate the forces estimation and support our theoretical statements.

 % Note that both $\sigma$ and $\tau$ are vital for forces estimation. To avoid tuning two parameters at the same time, 
 To evaluate the accuracy of the force targets in denoising pre-training, we quantify the precision by using the Pearson correlation coefficient $\rho$ between the estimated forces and ground truth. The force target estimation method is elucidated in detail in Supplementary Information \ref{app: setting}. $C_{error}$ denotes the estimation error. The ground truth forces are established by leveraging a supervised force learning method sGDML~\cite{chmiela2019sgdml}. For a fair comparison, we decouple the sampling and force calculation. The samples are drawn by perturbing the equilibrium conformation of molecule aspirin with noise setting ($\tau=0.04$), ($\sigma=1$, $\tau=0.04$), ($\sigma=20$, $\tau=0.04$), representing samples from near to far from the equilibrium. The results are shown in \ref{fig:md17&analysis3}.

Primarily, across all sampling settings, a hybrid of rotation noise and coordinate noise consistently outperforms the exclusive use of coordinate noise in terms of force accuracy. Specifically, the configuration of $\sigma=20, \tau=0.04$ demonstrates the optimal alignment with the ground truth force field. Considering $C_{error}$ tends to increase as $\sigma$ becomes larger, we choose $\sigma=2$, $\tau=0.04$ as the noise scale of Frad(RN). Furthermore, the accuracy gap between the hybrid noise and the coordinate noise is particularly evident when incorporating samples located farther from the equilibrium. As a consequence, the introduction of chemical-aware noise emerges as a pivotal strategy to enhance the accuracy of force learning targets.
% These two findings demonstrate the superiority of hybrid noise over coordinate noise. 3. % $C_{error}$ remains small in the settings with little angle noise scale, confirming the correctness of Proposition~\ref{prop1}. When $\sigma\leq 20$, $C_{error}$ grows large, indicating an inaccuracy force calculated by Table \ref{tab:ff}. % As a consequence, we choose $\sigma=2$, $\tau=0.04$ as the hyperparamenter of Frad.

\subsubsection{Frad can sample farther from the equilibriums}\label{sec:sampling coverage exp}

% This is because when using coordinate Gaussian noise (CGN) with isotropic noise variance, the noise level has to be set very low to prevent the generation of unreasonable substructures. However, under the low noise level, 
% They can only generate structures close to the equilibrium and are unable to cover common low-energy structures. Consequently, these methods have difficulty accurately learning the force for other common low-energy structures, which are essential for various downstream tasks.

% In contrast to existing coordinate denoising methods, Frad allows for a larger variance of T-rot noise, enabling the exploration of the energy landscape and covering more meaningful low-energy structures without generating invalid noisy structures. 
To compare the sampling coverage of different noise types, we measure it by perturbation scale, defined as the mean absolute coordinate changes of all atoms after applying noise. The perturbation scales, along with corresponding downstream performances of Coord and Frad measured by MAE, are depicted in \ref{fig:md17&analysis4}. 
The findings are threefold. Firstly, the challenge of low sampling coverage is evident in Coord. Specifically, the downstream performance is sensitive to the variance of CGN, where $\tau=0.04$ behaves best and both larger and smaller noise scales degenerate the performance significantly. This phenomenon can be attributed to larger noise scales yielding more irrational noisy samples, while smaller scales result in trivial denoising tasks. Such behavior aligns with the findings of \citet{SheheryarZaidi2022PretrainingVD}, who identified $\tau=0.04$ as the optimal hyperparameter. 
Secondly, rotation noise(RN) alleviates the low sampling coverage problem. Notably, the perturbation scale of RN can be increased significantly without losing competence. Even with $\sigma=20$, a setting with a perturbation scale 20 times larger than that of the most effective Coord configuration ($\tau=0.04$), Frad(RN) still outperforms Coord in all setups.
Thirdly, the comparison between ($\tau=0.04$) and ($\sigma=2$ or $ 1$, $\tau=0.04$) reveals that more accurate force approximations contribute to downstream task performance, because Frad gains further improvements over Coord despite sharing a comparable perturbation scale.

\section{Conclusion}\label{sec13}
In this work, we present a novel molecular pre-training framework, namely Frad, to learn effective molecular representations. Specifically, we propose a fractional denoising method coupled with a hybrid noise strategy to guarantee a force learning interpretation and meanwhile enable flexible noise design. We incorporate chemical priors to design chemical-aware noise and achieve a more refined molecular distribution modeling. Thus Frad can sample farther low-energy conformations from the equilibrium and learn more accurate forces of the sampled structures. As a result, Frad outperforms pre-training and non-pre-training baselines on force prediction, quantum chemical property prediction, and binding affinity prediction tasks. Besides, we showcase the robustness of Frad to inaccurate 3D data. We also validate that Frad surpasses coordinate denoising through improved force accuracy and enlarged sampling coverage. 
% We finally analyse the necessity of Frad strategy.

Our work offers several promising avenues for future exploration. Firstly, augmenting the volume of pre-training data has the potential to significantly enhance overall performance. The currently employed pre-training dataset is still much smaller than 2D and 1D molecular datasets due to the high cost of obtaining accurate molecular conformations. We anticipate more 3D molecular data to be available. Secondly, our current focus lies in property prediction using 3D inputs. By integrating with other pre-training methods, a model that can handle molecular tasks across data modalities may be produced. Ultimately, how to design chemical-aware noise for typical categories of molecules is worth investigation, such as nucleic acids, proteins and materials, so that Frad can be efficiently applied to a broader range of fields and expedite drug and material discovery. \added{These advancements hold the potential to establish Frad as a strong molecular foundation model applicable to diverse molecular tasks. Such progress could catalyze breakthroughs in critical areas like drug discovery and material science.}
%本文已经在Frad_ICML上探索了：Frad噪声能否推广到一般性的两阶段噪声？Frad能否推广到更多下游应用？Frad是否能用于噪声数据从而有潜力使用更多数据？
\section{Methods}\label{sec11}
\subsection{Model details}
In Frad pre-training, the equilibrium molecules extracted from the pre-training dataset are pre-processed by adding random hybrid noise. Then the noisy 3D molecules are encoded by a graph-based equivariant Transformer denoted as ${\rm Encoder}$. A denoising head ${\rm NoiseHead}$ made up of MLPs is adopted to predict the coordinate Gaussian noise (CGN) noise from the encoded features. The pre-training objective is given by
\begin{equation*}
    \mathcal{L}_{Frad}=\mathbb{E}_{x_{eq}}  \mathbb{E}_{p (x_{fin}|x_{med})p(x_{med}|x_{eq})p(x_{eq})}||{\rm NoiseHead}({\rm Encoder}(x_{fin})) - (x_{fin}-x_{med})||_2^2 ,
\end{equation*}
  where $x_{eq}$, $x_{med}$, $x_{fin}$ are equilibrium conformations, intermediate and final noisy conformations respectively. 
  During fine-tuning, the model is initialized from pre-training and property prediction heads ${\rm PropHead}$ specified for each downstream tasks are further optimized. The fine-tuning objective is given by
  \begin{equation*}
    \mathcal{L}_{FT}=\mathbb{E}_x||{\rm PropHead}({\rm Encoder}(x)) - {\rm Label} (x)||_{2}^{2}.
\end{equation*}
   Additionally, during fine-tuning, we also utilize Noisy Nodes techniques to further improve the performance. We proposed ``Frad Noisy Nodes" for tasks that are sensitive to input conformations such as MD17. Detailed algorithms and results of an ablation study are provided in Supplementary Information~\ref{app sec:algorithms} and \ref{section:exp NN}.

For the equivariant Transformer, it \added{primarily} follows the structure of TorchMD-NET~\cite{tholke2022torchmd}. \added{A model illustration is exhibited in \ref{fig:architecture}. We make some minor modifications to TorchMD-NET marked in dotted orange boxes in the figure: Firstly, to stabilize training, we add an additional normalization module in the residue updating, which is effective for both the QM9 and LBA tasks. Secondly, for the LBA task, due to the complexity of the input protein-ligand complex, we enhance the model's expressivity by incorporating angular information into the molecular geometry embedding used in the attention module. }

The model contains an embedding layer and multiple update layers. In the embedding layer, atom types are encoded by an atom-intrinsic embedding and a neighborhood embedding. These embeddings are then combined to generate the ultimate atomic embedding, serving as the initialization for the scalar feature in update layers.
The update layer includes attention-based interatomic interactions and information exchange to update the invariant scalar features $u_i$ and the equivariant vector features $\mathbf{v}_i$, where $i$ represents the atom index. The attention mechanism integrates \replaced{geometry}{edge} embeddings, derived from interatomic distances\replaced{, bond angles and torsions}{ $r_{ij}$} and filtered by \deleted{radial} basis functions\deleted{ denoted by $e^{RBF}$}. \replaced{After passed through one-layer MLP, the edge embeddings are denoted as $D_l, l=1,2$.}{The edge embeddings pass through one-layer MLP: $D_l=\sigma({\rm linear}( e^{RBF}(r_{ij}))), l=1,2$.} The attention weight matrix is defined by   
\begin{equation*}
    A_{ij}={\rm SiLU}({\rm dot}(W_Q u_i, W_K u_j,D_1)) \cdot \phi(r_{ij} ),
\end{equation*}
where $W_Q$ and $W_K$ are learnable weight matrices, ${\rm dot}$ refers to the extended dot product, i.e. an elementwise multiplication and subsequent sum over the feature dimension, $\phi$ is a cosine cutoff function, guaranteeing that atoms do not interact with a distance larger than the cutoff threshold.

The scalar feature $u$, along with $D_2$, is employed to generate intermediate update values, denoted as $s$, where $s_{ij}^1,s_{ij}^2,s_{ij}^3={\rm split}(W_V u_j \odot D_2)$. Subsequently, $s^3$ is treated as attention values for element-wise multiplication with matrix $A$, resulting in updated feature  $y_i={\rm linear}(\sum_j A_{ij}s_{ij}^3)$. These $y$ and $s$  values are then utilized to compute scalar updates $\Delta{u}$, and vector feature updates $\Delta{\bold{v}}$, as outlined in the following equations:

\begin{equation*}
q_{i}^1,q_{i}^2,q_{i}^3={\rm split}(y_i)
\end{equation*}

\begin{equation*}
    \Delta{u_i}=q_{i}^{1} + q_{i}^{2} \odot \left \langle \rm linear(\bold{v}_i), \rm linear(\bold{v}_i) \right \rangle,
\end{equation*}

\begin{equation*}
    \Delta{\bold{v}}_i= \sum^j_N{\left( {s_{ij}^1 \odot \bold{v}_j   + s_{ij}^2 \odot \vec{r}_{ij} } \right)} + q_{i}^{3} \odot \rm \rm linear(\bold{v}_i)
\end{equation*}
where $\left\langle \cdot , \cdot \right\rangle$ denotes the scalar product operation, and $\vec{r}_{ij}$ represents the normalized directional vector between nodes $x_i$ and $x_j$. $\Delta{u_i}$ and $\Delta{\bold{v}}_i$ are used as residual additional item to update $u_i$ and $\bold{v}_i$, respectively. \added{We further perform layer norm on the updated $u_i$ and $\bold{v}_i$ and use them as the output of the update layer.}
% The attention values also incorporates interatomic distance information and is split into three scalar representations: $s_{ij}^1,s_{ij}^2,s_{ij}^3={\rm split}(W_V x_j \odot D_2)$, yielding updated feature vectors $y_i={\rm linear}(\sum_j A_{ij}s_{ij}^3)$. The vector features are updated by incorporating directional vectors between two atoms and interacting with scalar features. Finally, the invariant scalar features and the equivariant vector features are updated.

\subsection{Dataset}\label{sec:app dataset}
All datasets used in pre-training and fine-tuning are listed in Table~\ref{table:datasets}.
% \vspace{-15pt}
\begin{table}[t]
\setlength{\tabcolsep}{3pt}
    \caption{Datasets statistics description.}
    \label{table:datasets}
    \begin{center}
    \begin{footnotesize}
    \begin{tabular}{lrrrrr}
    \toprule
    	Dataset & $\#$molecules	& 	$\#$conformations		& $\#$elements & $\#$atoms & Molecule types
     \\
    \midrule
    PCQM4Mv2 &3378606& 1 per molecule & 22 &  $\approx$30  & Stable molecules\\
    QM9 & 133,885& 1 per molecule &5 &  3-29 & Stable small organic molecules\\
     ISO17 & 129 &5000 per molecule &3 & 19 & Isomers of $C_7 O_2H_{10}$\\
     MD17 & 8 & 3611115 in total  &4 & 9$-$21 & Small organic molecules\\
     MD22 &7 &223422 in total &4  &42-370 &  \makecell[l]{Proteins, lipids, carbohydrates,\\ nucleic acids, supramolecules}\\
     % ANI$-$1x  &63865 &5496771 & 4 &2$-$63 & Small organic molecules\\
     LBA (Atom3D) & 4,598 complexes& 1 per complex  & $\approx$10 & $<600$ & Protein-ligand complexes\\
    %  PCQM4Mv2 &\multicolumn{2}{c}{\makecell[c]{3378606 molecules,\\ each has one conformation}} & 22 &  $\approx$30  & stable molecules\\
    % QM9 & \multicolumn{2}{c}{\makecell[c]{133,885 molecules, \\each has one conformation}} &5 &  3-29 & stable small organic molecules\\
    %  ISO17 & 129 &645000 &3 & 19 & Isomers of $C_7 O_2H_{10}$\\
    %  MD17 & 8 & 3611115  &4 & 9$-$21 & Small organic molecules\\
    %  MD22 &7 &223422 &4  &42-370 &  \makecell[l]{Proteins, lipids, carbohydrates,\\ nucleic acids, supramolecules}\\
    %  % ANI$-$1x  &63865 &5496771 & 4 &2$-$63 & Small organic molecules\\
    %  LBA (Atom3D) & \multicolumn{2}{c}{\makecell[c]{4,598 complexes, \\each has one conformation}}  & $\approx$10 & $<600$ & protein-ligand complexes\\
    \bottomrule
    \end{tabular}
    \end{footnotesize}
    \end{center}
    \vskip -0.1in
\end{table}

%C4H4N2O2 12,C6H6 12, C10H8 18,C9H8O4  21, C7H6O3 16,C3H4O2 9, C2H6O 9 ,C7H8 15
% 'aspirin', 'benzene2017', 'ethanol', 'malonaldehyde', 'naphthalene', 'salicylic', 'toluene', 'uracil': 211762+ 627983+555092+ 993237+326250+320231+442790+133770=3,611,115

% We %数据介绍列表 split按照jctc来:80%/10%/10%into the train,validation,and test sets for ANI-1, ANI-1x,MD22;ISO17:each trajectory consisting of 5,000 conformations, We draw a subset of 4,000 steps from 80% of the MD trajectories for training and validation. This leaves us with a separate test set for each scenario: (1) the unseen 1,000 conformations of molecule trajectories included in the training set and(2) all 5,000 conformations of the remaining 20% of molecules not included in training

We utilize PCQM4Mv2~\cite{nakata2017pubchemqc} as the pre-training dataset. It contains 3.4 million organic molecules, with one equilibrium conformation and one label calculated by density functional theory (DFT). We do not use the label since our method is self-supervised. 

We adopt three different kinds of downstream tasks. \added[id=2]{The splitting methods are different for different tasks to maintain chemical diversity or to test out of-distribution generalization. The splittings all adhere to standard practices in the literature to ensure fair comparisons.}

For atom-wise force prediction, we utilized molecular dynamics trajectories dataset MD17~\cite{chmiela2017machine}, MD22~\cite{Chmiela2023MD22} and ISO17~\cite{schutt2017schnetiso17}. 

MD17 is a dataset of molecular dynamics trajectories, containing 8 small organic molecules with conformations, total energy, and force labels computed by the electronic structure method. For each molecule, 150k to nearly 1M conformations are provided.
For MD17 data splitting, the existing approaches \added[id=2]{all perform uniformly random splitting, but} diverge on taking large (9500) or small (950 or 1000) size of training data. As the size of the training dataset affects the force prediction significantly, we perform Frad with both splitting for fair comparisons.

Compared to MD17, MD22 presents new challenges with respect to system size, flexibility, and degree of nonlocality, containing molecular dynamics trajectories of proteins, carbohydrates, nucleic acids, and supramolecules.
ISO17 consists of molecular dynamics trajectories of 129 isomers of C7O2H10. While the composition of all included molecules is the same, the chemical structures are fundamentally different. 
To fairly compare with baseline Coord(Noneq)~\citet{coordnoneq2023}, we follow their data splitting settings for MD22 and ISO17. For MD22, the dataset is \added[id=2]{randomly} split by a ratio of $80\%/10\%/10\%$ into the train, validation, and test sets. For ISO17, the splitting strategy follows the original literature~\cite{schutt2017schnetiso17}, where\deleted{conformations of} the $80\%$ of molecules are used for training and validation, and \replaced{conformations of those unseen $20\%$ of molecules are used for testing. Thus the test sets cover a different chemical space from the training and validation sets, making the task more challenging and aligning well with the real-world scenarios.}{all conformations of the remaining $20\%$ of molecules are used for testing.}

For molecule-wise quantum chemical property prediction, we use the QM9 dataset\cite{ruddigkeit2012enumeration,ramakrishnan2014quantum}. 
QM9 is a quantum chemistry dataset including geometric, energetic, electronic, and thermodynamic properties for 134k stable small organic molecules made up of CHONF atoms. Each molecule has one equilibrium conformation and 12 labels calculated by density functional theory (DFT). The QM9 dataset is \added[id=2]{randomly} split into a training set with 110,000 and a validation set with 10,000 samples, leaving 10,831 samples for testing. This splitting is commonly applied in the literature. As usually done on QM9, we fine-tune a separate model for each of the 12 downstream tasks, with the same pre-trained model.

For complex-wise binding affinity prediction, we adopt the LBA dataset in Atom3D benchmark~\cite{atom3d}, containing conformations of protein-ligand complexes and their corresponding binding strengths. Splitting follows Atom3D such that no protein in the test dataset has more than $30\%$ or $60\%$ sequence identity with any protein in the training dataset.

\subsection{Baselines}
The baselines include pre-training methods and non-pre-training methods. In terms of 3D pre-training approaches, our baselines cover the currently SOTA methods we have known, including Coord~\cite{SheheryarZaidi2022PretrainingVD}, 3D-EMGP~\cite{jiao2022energy}, SE(3)-DDM~\cite{ShengchaoLiu2022MolecularGP}, Transformer-M~\cite{luo2022one}. Coord is the baseline we are most interested in because it is a typical coordinate denoising pre-training method and shares the same backbone with Frad. So their performance well reflects the comparison between coordinate denoising and fractional denoising, reflecting the effect of molecular distribution modeling. 3D-EMGP and SE(3)-DDM are denoising methods based on coordinate denoising but with an invariant energy function, a symmetric constraint for 3D molecular representation. As for Transformer-M, it is a competitive model consisting of denoising and energy prediction pre-training tasks. Coord(NonEq) pre-train by coordinate denoising on large nonequilibrium datasets ANI-1 and ANI-1x.
 % We exclude Uni-mol~\cite{zhou2023unimol} and ChemRL-GEM \cite{fang2021chemrl} since they only provide the average performance of 3 energy tasks in QM9. 
 
We also adopt the representative approaches designed for property prediction tasks without pre-training. The models for quantum chemical property prediction comprise TorchMD-NET~\cite{tholke2022torchmd}, SchNet~\cite{schutt2018schnet}, E\deleted{(n-)}GNN\cite{satorras2021n}, DimeNet~\cite{gasteiger2020directional}, DimeNet++~\cite{klicpera2020fast}, SphereNet~\cite{liu2022spherical}, PaiNN~\cite{schutt2021equivariant}. As for binding affinity prediction, the baselines including sequence-based methods such as DeepDTA~\cite{ozturk2018deepdta}, TAPE~\cite{rao2019TAPE}, and ProtTrans~\cite{elnaggar2021prottrans} and structure-based methods, such as various variants of Atom3D~\cite{atom3d}, Holoprot~\cite{somnath2021Holoprot}, and ProNet~\cite{wang2022PRONET}.

We use the publicly available code of the original paper to produce results for Coord and Coord(NonEq) with the same backbone model as ours. The result of TorchMD-NET with 9500 training data of MD17 is reported by~\cite{jiao2022energy}. Other results are taken from the referred papers.

% Topical subheadings are allowed. Authors must ensure that their Methods section includes adequate experimental and characterization data necessary for others in the field to reproduce their work. Authors are encouraged to include RIIDs where appropriate. 

% \textbf{Ethical approval declarations} (only required where applicable) Any article reporting experiment/s carried out on (i)~live vertebrate (or higher invertebrates), (ii)~humans or (iii)~human samples must include an unambiguous statement within the methods section that meets the following requirements: 
% \begin{enumerate}[1.]
% \item Approval: a statement which confirms that all experimental protocols were approved by a named institutional and/or licensing committee. Please identify the approving body in the methods section
% \item Accordance: a statement explicitly saying that the methods were carried out in accordance with the relevant guidelines and regulations
% \item Informed consent (for experiments involving humans or human tissue samples): include a statement confirming that informed consent was obtained from all participants and/or their legal guardian/s
% \end{enumerate}
% If your manuscript includes potentially identifying patient/participant information, or if it describes human transplantation research, or if it reports results of a clinical trial then  additional information will be required. Please visit (\url{https://www.nature.com/nature-research/editorial-policies}) for Nature Portfolio journals, (\url{https://www.springer.com/gp/authors-editors/journal-author/journal-author-helpdesk/publishing-ethics/14214}) for Springer Nature journals, or (\url{https://www.biomedcentral.com/getpublished/editorial-policies\#ethics+and+consent}) for BMC.

\section*{Data Availability}
    The pre-training and fine-tuning data used in this work are available in the following links:
    
    \added{ 
    PCQM4Mv2~\cite{nakata2017pubchemqc}: \url{https://ogb.stanford.edu/docs/lsc/pcqm4mv2/}, \url{https://figshare.com/articles/dataset/MOL_LMDB/24961485}; }
    
    \added{ QM9~\cite{ruddigkeit2012enumeration,ramakrishnan2014quantum}: \url{https://figshare.com/collections/Quantum_chemistry_structures_and_properties_of_134_kilo_molecules/978904};} 
    
    \added{MD17~\cite{chmiela2017machine} and MD22~\cite{Chmiela2023MD22}: \url{http://www.sgdml.org/\#datasets}; }
    
    \added{ ISO17~\cite{schutt2017schnetiso17}: \url{http://quantum-machine.org/datasets/}; }
    
    \added{ LBA~\cite{atom3d}: \url{https://zenodo.org/records/4914718}.}
    
    \deleted{PCQM4Mv2:\url{http://ogb-data.stanford.edu/data/lsc/pcqm4m-v2-train.sdf.tar.gz},\url{https://figshare.com/articles/dataset/MOL_LMDB/24961485}; QM9:\url{https://ndownloader.figshare.com/files/3195404}; }\deleted{MD17:\url{http://quantum-machine.org/gdml/data/npz}; 
    MD22:\url{http://www.sgdml.org/\#datasets}; ISO17: \url{http://quantum-machine.org/datasets/iso17.tar.gz}; 
    LBA:\url{https://zenodo.org/records/4914718}.}
    
    The source data for the figures in this work~\added[id=2]{\cite{Ni2024}} are available in the following link: \url{https://doi.org/10.6084/m9.figshare.25902679.v1}
    
\section*{Code Availability} 

    Source codes for Frad pre-training and fine-tuning are available at \url{https://github.com/fengshikun/FradNMI}. The pre-trained models \cite{feng_2024_12697467} are available at \url{https://zenodo.org/records/12697467}

\backmatter

% \bmhead{Supplementary information}
% Supplementary information including supplementary theoretical analysis, theoretical proofs, supplementary experiments, experimental settings, and related work is included in the Supplementary Information.
% If your article has accompanying supplementary file/s please state so here. 

% Authors reporting data from electrophoretic gels and blots should supply the full unprocessed scans for key as part of their Supplementary information. This may be requested by the editorial team/s if it is missing.

% Please refer to Journal-level guidance for any specific requirements.

% \bmhead{Acknowledgments}

\section*{Acknowledgements}
    Y.L. acknowledges funding from National Key R\&D Program of China No.2021YFF1201600 and Beijing Academy of Artificial Intelligence (BAAI).

\section*{Author Contributions Statement}
     Y.N. and S.F. conceived the initial idea for the projects. S.F. processed the dataset and trained the model. Y.N. developed the theoretical results and drafted the initial manuscript. X.H. and Y.N. analyzed the results and created illustrations and data visualizations. S.F. and Y.S. carried out the experiments utilizing the pre-trained model. \replaced[id=2]{Y.N., S.F., X.H., and Y.L. participated in the revision of the manuscript. The project was supervised by Y.L., Q.Y., Z.M., and W.M., with funding provided by Y.L. and Q.Y.}{The revision of the manuscript involved contributions from all authors. The project was under the supervision of Y.L. with active participation from all authors in discussions.}

\section*{Competing Interests Statement}
    The authors declare no competing interests.

% Acknowledgments are not compulsory. Where included they should be brief. Grant or contribution numbers may be acknowledged.

% Please refer to Journal-level guidance for any specific requirements.
\captionsetup[figure]{labelformat=empty}
\renewcommand{\thefigure}{Extended Data Fig.\arabic{figure}}
\setcounter{figure}{0}

\begin{figure}[h!]
\begin{center}
\includegraphics[width=0.92\textwidth]{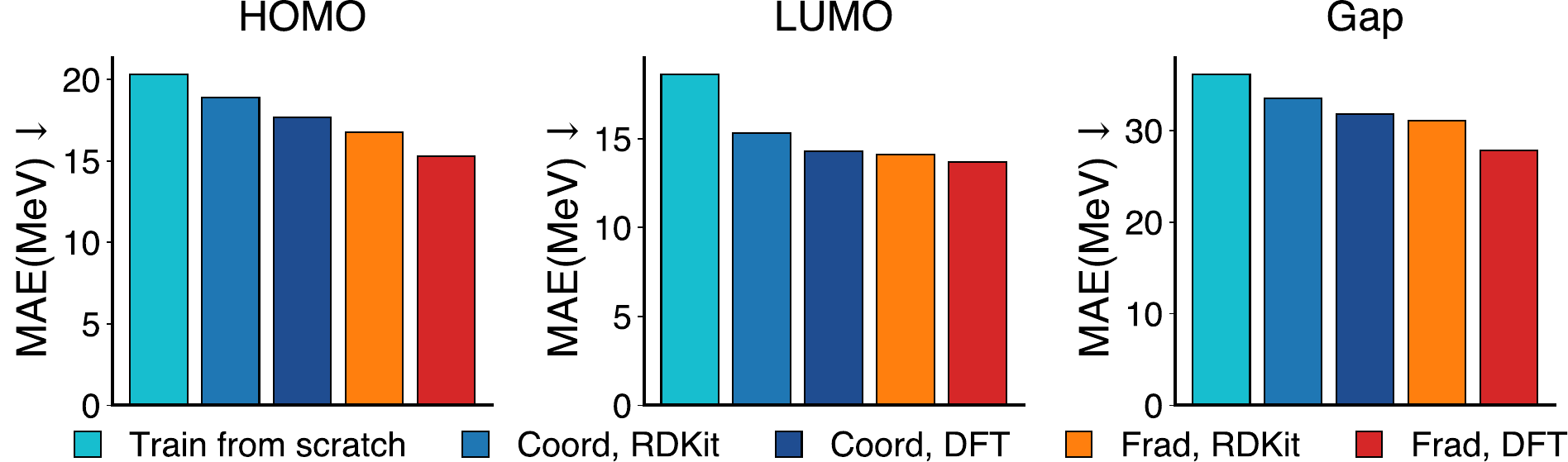}

\caption{\textbf{\thefigure:} \textbf{Test performance of Coord and Frad with different data accuracy on 3 tasks in QM9.} "Train from scratch" refers to the backbone model TorchMD-NET without pre-training. Both Coord and Frad use the TorchMD-NET backbone. "RDKit" and "DFT" refer to pre-training on molecular conformations generated by RDKit and DFT methods respectively. We can see that Frad is more robust to inaccurate pre-training conformations than Coord.} 
 \label{fig:md17&analysis2}
 \end{center}
\end{figure}

\begin{figure}[h!]
\begin{center}
\includegraphics[width=1\textwidth]{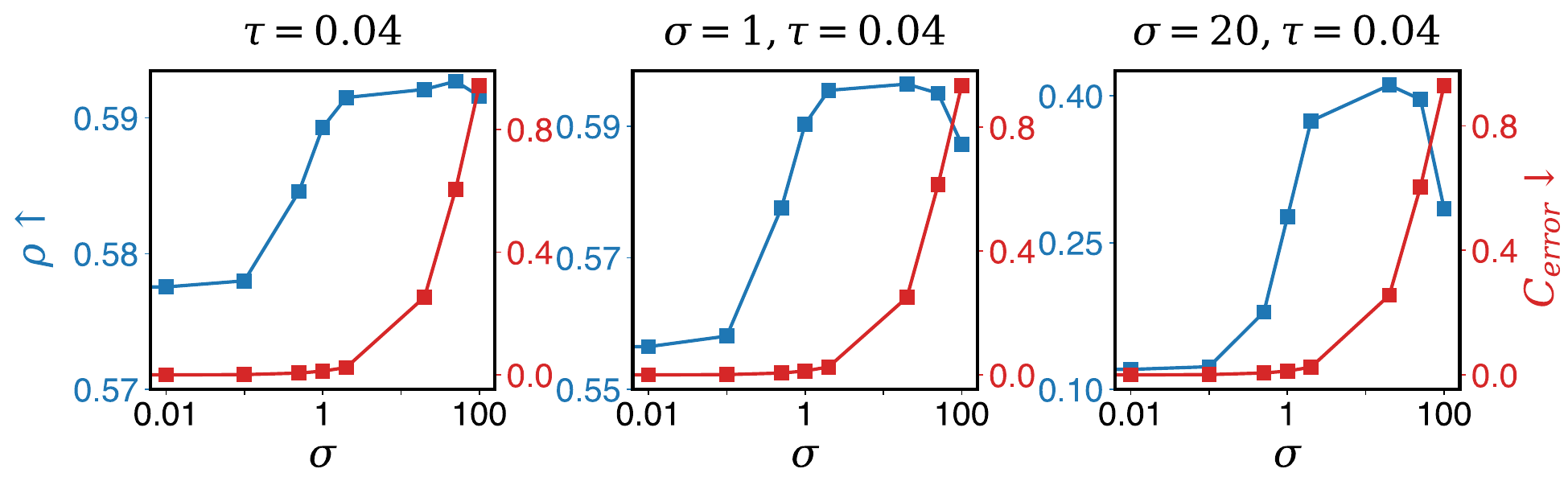}
\caption{\textbf{\thefigure:} \textbf{Estimated force accuracy of Frad and Coord in three different sampling settings.} Force accuracy is measured by Pearson correlation coefficient $\rho$ between the force estimation and ground truth. The estimation error is denoted as $C_{error}$. We can see that within the range of small estimation errors, the force estimated by Frad($\sigma>0$) is consistently more accurate than that estimated by Coord($\sigma=0$).}
 \label{fig:md17&analysis3}
 \end{center}
\end{figure}

\begin{figure}[h!]
\begin{center}
\includegraphics[width=0.5\textwidth]{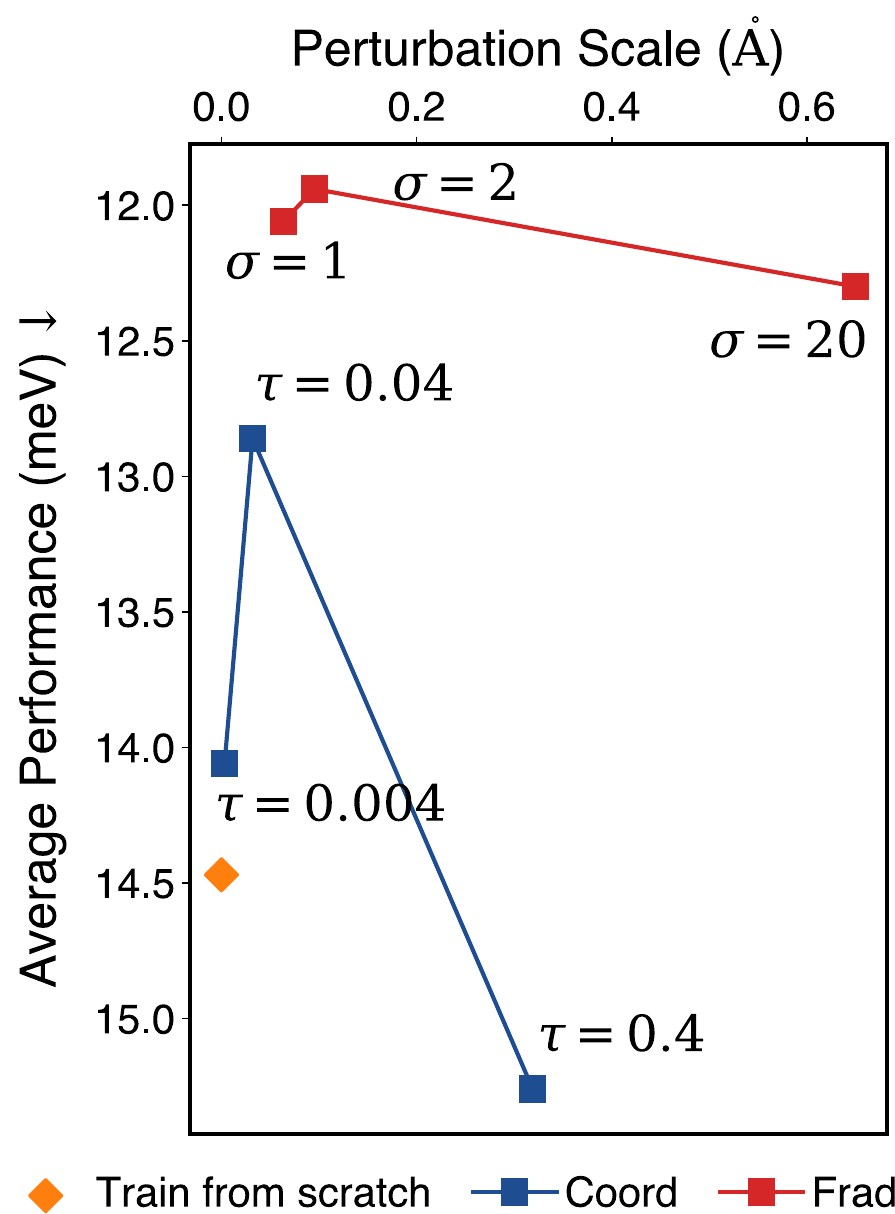}
\caption{\textbf{\thefigure:} \textbf{Performance of Coord and Frad with different perturbation scales.} The performance (MAE) is averaged across seven energy prediction tasks in QM9. The perturbation scale refers to the mean absolute coordinate changes resulting from noise application and is determined by the noise scale. \added{For Frad, the hyperparameter is fixed at $\tau=0.04$.} We can see that Frad can effectively sample farther from the equilibrium, \added{without a significant performance drop.}}
 \label{fig:md17&analysis4}
 \end{center}
\end{figure}

\newpage
% \noindent
% If any of the sections are not relevant to your manuscript, please include the heading and write `Not applicable' for that section. 

\bigskip

% Springer journals and proceedings: \url{https://www.springer.com/gp/editorial-policies}

% Nature Portfolio journals: \url{https://www.nature.com/nature-research/editorial-policies}

% \textit{Scientific Reports}: \url{https://www.nature.com/srep/journal-policies/editorial-policies}

% BMC journals: \url{https://www.biomedcentral.com/getpublished/editorial-policies}
% \end{flushleft}
\bibliography{main}% common bib file

\newpage
\begin{appendix}

\captionsetup[figure]{labelformat=empty}
\captionsetup[table]{labelformat=empty}
\renewcommand{\thefigure}{Supplementary Figure \arabic{figure}}
\renewcommand{\thetable}{Supplementary Table \arabic{table}}

\begin{center}
    \section*{\Large Supplementary Information}
\end{center}
\vspace{2pt}
\section{Supplementary theoretical analysis}
\subsection{Challenges of coordinate denoising}\label{app:Chanllenges of coordinate denoising}
In this section, we first provide the proof of physical interpretation for coordinate denoising, and then discuss its challenge of sampling and force accuracy. To simplify notations, we substitute $x_{f}$ for $x_{fin}$ to denote the final noisy conformation.
\begin{theorem}[Learning Forces via Coordinate Denoising~\cite{SheheryarZaidi2022PretrainingVD}]\label{thm:coord}
If the distribution of coordinate noise is isotropic Gaussian $p(x_{f}|x_{eq})\sim \mathcal{N}(x_{eq},\tau^2  I_{3N})$, then coordinate denoising is equivalent to learning the atomic forces that correspond to the approximate molecular distribution by Boltzmann Distribution. 
 % \vspace{-5pt}
     \begin{small} 
    \begin{subequations}
     % \vspace{-3pt}
    \begin{equation}
        \mathbb{E}_{p (x_f|x_{eq})p(x_{eq})}||GNN_{\theta} (x_f) - (x_f-x_{eq})||^2 \label{eq:coord target in thm}
    \end{equation}
     \begin{equation}
     \begin{aligned}
          \simeq  \mathbb{E}_{p (x_f)}||GNN_{\theta} (x_f) - \nabla _{x_f} E(x_f)||^2, \label{eq:learn ff target in coord thm}
     \end{aligned}
    \end{equation}      
    \end{subequations}
    \end{small} 
$\simeq$ denotes the equivalence as optimization objectives. \eqref{eq:coord target in thm} is the optimization objective of coordinate denoising~\cite{SheheryarZaidi2022PretrainingVD,luo2022one,feng2023may}. \eqref{eq:learn ff target in coord thm} is the regression task of fitting an approximate force field. 
\end{theorem}
We make some adjustments to the form of the proof provided by~\citet{SheheryarZaidi2022PretrainingVD}.
\begin{proof}
Define $\widetilde{GNN}_{\theta} (x_f)$ as $\frac{-1}{\tau^2}\cdot GNN_{\theta} (x_f)$ and
$\widetilde{\widetilde{GNN}}_{\theta} (x_f)$ as $kT\cdot\widetilde{GNN}_{\theta} (x_f)$. 
\begin{small}
 \begin{subequations}
\begin{align}
       LHS=&\mathbb{E}_{p (x_f,x_{eq})}\tau^4||\widetilde{GNN}_{\theta} (x_f) - \frac{-1}{\tau^2}(x_f-x_{eq}))||^2 \label{eq:thm pf coord a}\\
       =&  \mathbb{E}_{p (x_f,x_{eq})}\tau^4||\mbox{$\widetilde{GNN}$\added{$\theta$}} (x_f) - \nabla _{x_f} \log p (x_f|x_{eq})||^2 \label{eq:thm pf coord b}\\
       \simeq& \mathbb{E}_{p (x_f)}\tau^4||\mbox{$\widetilde{GNN}$\added{$\theta$}} (x_f) - \nabla _{x_f} \log p (x_f)||^2 +T_5\label{eq:thm pf coord c}\\
        =&\mathbb{E}_{p (x_f)}\tau^4||\widetilde{GNN}_{\theta} (x_f) -  (-\nabla _{x_f} \frac{E(x_f)}{kT})||^2 +T_5\label{eq:thm pf coord d}\\  
    =&\mathbb{E}_{p (x_f)}\frac{\tau^4}{(kT)^2}||\widetilde{\widetilde{GNN}}_{\theta} (x_f) -  (-\nabla _{x_f} E(x_f))||^2 +T_5 \simeq RHS  \label{eq:thm pf coord e} 
    \end{align}
    \end{subequations}        
\end{small}
    The equality between \eqref{eq:thm pf coord a} and \eqref{eq:thm pf coord b} uses the Gaussian distribution and the mixture model, i.e. $p (x_f, x_{eq})=p (x_f|x_{eq})p(x_{eq})$.
    The next equality between \eqref{eq:thm pf coord b} and \eqref{eq:thm pf coord c} uses the result in \cite{PascalVincent2011ACB}, which is also provided in Proposition \ref{app_prop:vincent}. $T_5$ represents terms do not contain $\theta$.
    The next equality between \eqref{eq:thm pf coord c} and \eqref{eq:thm pf coord d} applies the Boltzmann Distribution in Assumption \ref{assump:Boltzmann_Dist}. 
    The equivalence in \eqref{eq:thm pf coord e} holds because the constant factors $T_5$ and $\frac{\tau^4}{(kT)^2}$ do not affect the optimization for $\theta$. Besides, $\widetilde{\widetilde{GNN}}_{\theta}$ and $GNN_{\theta}$ are up to a constant factor, which is viewed as equivalence in \cite{SheheryarZaidi2022PretrainingVD}. Another understanding is that the two GNNs learn the same force field label but up to unit conversion.
\end{proof}
From the proof, we obtain an equivalent learning target for coordinate denoising as follows.
\begin{equation}\label{coord challenge}
     \begin{aligned}
          \mathcal{L}_{Denoising} \simeq 
          &\mathbb{E}_{p (x_{f})}||GNN_{\theta} (x_{f}) - (-\nabla _{x_{f}} \log p(x_{f}))||^2, 
     \end{aligned}
    \end{equation}   
where $(-\nabla _{x_{f}} \log p(x_{f}))=\nabla _{x_{f}} E(x_{f})$ indicate the force learning targets. In \eqref{coord challenge}, the learning target is affected by the modeled molecular distribution $p (x_{f})$ in two aspects, i.e. sampling and force target. As a result, the molecular distribution described by isotropic Gaussian causes two challenges to coordinate denoising:
\begin{itemize}
    \item \textbf{Low Sampling Coverage.} Existing coordinate denoising approaches often set a very small noise level to prevent the generation of distorted substructures such as distorted aromatic rings. Experimental observations, as reported by \citet{SheheryarZaidi2022PretrainingVD}, indicate a significant drop in performance when the noise level is increased. A similar phenomenon is discussed in section~\ref{sec:sampling coverage exp}. While the small noise level strategy effectively avoids rare and undesired noisy structures, it struggles to encompass prevalent structures with low energy, leading to low efficiency of force learning. Consequently, current coordinate denoising methods are constrained in learning forces for common low-energy structures far from the provided equilibriums.
    \item \textbf{Inaccurate Forces.} Existing coordinate denoising methods assume noise with an isotropic covariance, implying a uniform slope of the energy function in all directions around a local minimum. However, the intrinsic nature of a molecule's energy function is anisotropic. As depicted in Figure \ref{fig:overview}a, in low energy conformations, some substructures are rigid while some single bonds can rotate in a large scale. Consequently, current methods deviate from the true energy landscape, resulting in inaccurate force learning targets.

\end{itemize}

\subsection{Non-fractional denoising fails to hold the force learning interpretation}\label{sec:neccessity} %放附录？
Frad is a distinctive denoising framework that solely denoises a part of the noise introduced, presenting a departure from conventional methods which seek to predict the entire noise introduced and recover the original conformations. While denoising the complete hybrid noise seems like a feasible strategy, our analysis in this section reveals that it does not preserve the force learning interpretation. This limitation highlights the necessity and significance of the Frad approach.

In establishing the equivalence in theorem~\ref{thm:frad}, a crucial step is that the conditional score function is proportional to the coordinate changes. It is a property naturally satisfied by CGN. To investigate whether denoising CAN or the whole hybrid noise is meaningful, we consider CAN in the asymptotic case, i.e. when the noise scale approaches zero, so that the conditional score function is approximately linearly related to the coordinate changes. 
\begin{table}[h!]
\centering
\small
    \caption{\textbf{\thetable: The conformation distribution and conditional score function corresponding to various noise types.} The covariances $\Gamma_1(x_{eq})=C(x_{eq})\Sigma C(x_{eq})^\top$ and $\Gamma_2(x_{eq})=\tau^2 I_{3N}+C(x_{eq})\Sigma C(x_{eq})^\top$ are dependent on the input structure $x_{eq}$, indicating denoising the
complete hybrid noise or chemical-aware noise fails to hold the force learning interpretation and highlights the necessity of Frad. Detailed forms of $C$ and $\Sigma$ are provided in Supplementary Information~\ref{section:appendix proof table}.}
    \label{tab:ff}
     \vskip 0.15in
      % \begin{center}
      % \begin{small}
    % \begin{sc}
        % \begin{scriptsize}
        \begin{tabular}{l c c}
        \toprule 
          \makecell[c]{\textbf{Noise Type}} & \makecell[c]{  \textbf{Noise} \textbf{Distribution }\\$p(\cdot|x_{eq})$} & \makecell[c]{  \textbf{Conditional Score Function} \\ $\nabla \log p(\cdot|x_{eq})$ }\\
          \midrule
            \makecell[c]{CGN} &  $x_{CGN}\sim\mathcal{N}\left(x_{eq},\tau^2 I_{3N}\right)$ & $-\frac{1}{\tau^2}(x_{CGN}-x_{eq})$  \\  
          % \hline
           \makecell[c]{CAN} &  $x_{med}\sim\mathcal{N}\left(x_{eq},\Gamma_1(x_{eq})\right)$ & $-\Gamma_1(x_{eq})^{-1}(x_{med}-x_{eq})$ \\ 
          % \hline
          \makecell[c]{Hybrid(CAN+CGN)} &   $x_{fin}\sim\mathcal{N}\left(x_{eq},\Gamma_2(x_{eq})\right)$ & $- \Gamma_2(x_{eq})^{-1}(x_{fin}-x_{eq})$ \\
          \bottomrule
        \end{tabular}
        % \end{scriptsize}
    % \end{sc}
     % \end{small}
     %  \end{center}
       % \vskip 0.1in
        % \vspace{-20pt}
\end{table}

The molecular conformation distributions and conditional score functions corresponding to different noise types are presented in \ref{tab:ff}.  In the asymptotic scenario, the conditional score function is linear to the coordinate changes. We find only in the case of CGN, the linear coefficient is a constant. This implies that denoising is equivalent to fitting the forces because a constant can be viewed as a unit convert and maintain the force learning interpretation~\cite{SheheryarZaidi2022PretrainingVD}. In contrast, in the case of other noise types, the linear coefficient is dependent on the input equilibrium structure $x_{eq}$. Consequently, denoising these types of noise is not equivalent to fitting the atomic forces. The proof for the results presented in the table is provided in Supplementary Information~\ref{section:appendix proof table}.

% The detailed calculations of the distributions can be found in appendix~\ref{section:appendix proof table}. Note that the linear results are asymptotic, i.e. holds exactly when the noise scale approaches zero. Nonetheless, these results are still sufficient to illustrate the issue, as the equivalence between denoising and fitting the force should hold for all noise scales. Besides, if the covariance is not invertible, the force can not be written explicitly, but the unequivalence still holds.
 % \vspace{-14pt}

\section{Theoretical proofs}\label{section:appendix proof}
To simplify notations, we substitute $x_{f}$ and $x_{m}$ for $x_{fin}$ and $x_{med}$, denoting the final noisy conformation and the intermediate conformation respectively in this section.
\subsection{Proofs for Theorem \ref{thm:frad}}\label{section:appendix proof thm1}
\textbf{Theorem \ref{thm:frad}} (Learning Forces via Fractional Denoising).
If the distribution of hybrid noise satisfies $p(x_{fin}|x_{med})\sim \mathcal{N}(x_{med},\tau^2  I_{3N})$ is a coordinate Gaussian noise (CGN), then fractional denoising is equivalent to learning the atomic forces that correspond to the approximate molecular distribution by Boltzmann Distribution. 
 % \vspace{-5pt}
     \begin{small} 
    \begin{subequations}
     % \vspace{-3pt}
    \begin{equation}
        \mathbb{E}_{p (x_{f}|x_{m})p(x_{m}|x_{eq})p(x_{eq})}||GNN_{\theta} (x_{f}) - (x_{f}-x_{m})||^2 \label{app eq:frad target in thm}
    \end{equation}
     \begin{equation}
     \begin{aligned}
          \simeq  \mathbb{E}_{p (x_{f})}||GNN_{\theta} (x_{f}) - \nabla _{x_{f}} E(x_{f})||^2, \label{eq:learn ff target in thm}
     \end{aligned}
    \end{equation}      
     % \vspace{-2pt}
    \end{subequations}
    \end{small} 
$\simeq$ denotes the equivalence as optimization objectives. \eqref{app eq:frad target in thm} is the optimization objective of Frad. \eqref{eq:learn ff target in thm} is the regression task of fitting an approximate force field. 

% \begin{theorem}[Fractional Denoising Score Matching]\label{app_Anisotropic denoising score matching}
% If $p(x_f|x_m)\sim \mathcal{N}(x_m,\tau^2  I_{3N})$ and $p(x_m|x_{eq})$ can be arbitrary distribution, we have
% \begin{equation}
% \begin{aligned}\label{eq:app_Anisotropic denoising score matching}
%     &E_{p (x_f|x_m)p(x_m|x_{eq})p(x_{eq})}||GNN_{\theta} (x_f) - (x_f-x_m)||^2 \\
%     \simeq & E_{p (x_f)}||GNN_{\theta} (x_f) - (-\nabla _{x_f}  E(x_f))||^2 ,
% \end{aligned}
% \end{equation}
% where $\simeq$ denotes the equivalence as optimization objectives, $- \nabla _{x_f} E(x_f)$ is the forces corresponding to the molecular distribution modeled by hybrid noise. 
% % and $C=\frac{-\tau^2}{kT}$ is a constant independent of $x_f$ and $\theta$. 
% \end{theorem}  
\begin{proof}
Define $\widetilde{GNN}_{\theta} (x_f)$ as $\frac{-1}{\tau^2}\cdot GNN_{\theta} (x_f)$ and
$\widetilde{\widetilde{GNN}}_{\theta} (x_f)$ as $kT\cdot\widetilde{GNN}_{\theta} (x_f)$. 
\begin{subequations}
\begin{align}
    LHS
    =&\mathbb{E}_{p (x_f|x_m)p(x_m|x_{eq})p(x_{eq})}\tau^4||\widetilde{GNN}_{\theta} (x_f) - \frac{-1}{\tau^2}(x_f-x_m))||^2 \label{eq:thm pf a}\\  
    =&\mathbb{E}_{p (x_f,x_m,x_{eq})}\tau^4||\widetilde{GNN}_{\theta} (x_f) - \nabla_{x_f} \log p (x_f|x_m)||^2 \label{eq:thm pf b}\\  
    =&\mathbb{E}_{p (x_f,x_m)}\tau^4||\widetilde{GNN}_{\theta} (x_f) - \nabla_{x_f} \log p (x_f|x_m)||^2 \label{eq:thm pf c}\\  
    =&\mathbb{E}_{p (x_f)}\tau^4||\widetilde{GNN}_{\theta} (x_f) - \nabla_{x_f} \log p (x_f)||^2 +T_1 \label{eq:thm pf d}\\  
    =&\mathbb{E}_{p (x_f)}\tau^4||\widetilde{GNN}_{\theta} (x_f) -  (-\nabla _{x_f} \frac{E(x_f)}{kT})||^2 +T_1 \label{eq:thm pf e}\\  
    =&\mathbb{E}_{p (x_f)}\frac{\tau^4}{(kT)^2}||\widetilde{\widetilde{GNN}}_{\theta} (x_f) -  (-\nabla _{x_f} E(x_f))||^2 +T_1\label{eq:thm pf f} 
    % =&E_{p (x_f)}\frac{\tau^4}{(kT)^2}||\frac{-kT}{\tau^2}GNN_{\theta} (x_f) -  \frac{-kT}{\tau^2}C(-\nabla _{x_f} E(x_f))||^2 +T_1\label{eq:thm pf g}\\  
    % =&E_{p (x_f)}||GNN_{\theta} (x_f) -  C(-\nabla _{x_f} E(x_f))||^2 +T_1   \label{eq:thm pf h} 
    \simeq RHS,
\end{align}
\end{subequations}
The equality between \eqref{eq:thm pf a} and \eqref{eq:thm pf b} uses the gaussian distribution condition $p(x_f|x_m)\sim \mathcal{N}(x_m,\tau^2  I_{3N})$ and the fact that the noise adding process $x_{eq}\rightarrow x_m\rightarrow x_f$ is a Markov chain, i.e. $p (x_f,x_m,x_{eq})=p (x_f|x_m)p(x_m|x_{eq})p(x_{eq})$.

The next equality between \eqref{eq:thm pf b} and \eqref{eq:thm pf c} holds because the term in the expectation, simply denoted as $F(x_f,x_m)$, does not contain $x_{eq}$. To be specific, the relationship between joint and marginal density functions is given by $p (x_f,x_m)=\int p (x_f,x_m,x_{eq}) \mathrm{d}x_{eq}$, thus with fubini theorem, we have $\mathbb{E}_{p (x_f,x_m,x_{eq})}F(x_f,x_m)$ $=\int\int\int p (x_f,x_m,x_{eq})F(x_f,x_m) \mathrm{d}x_f\mathrm{d}x_m\mathrm{d}x_{eq}$ $=\int\int(\int p (x_f,x_m,x_{eq})\mathrm{d}x_{eq}) F(x_f,x_m)\mathrm{d}x_f\mathrm{d}x_m$ $=\int\int p (x_f,x_m)F(x_f,x_m) \mathrm{d}x_f\mathrm{d}x_m$. 

The next equality between \eqref{eq:thm pf c} and \eqref{eq:thm pf d} uses the result in \cite{PascalVincent2011ACB}, which is also provided in Proposition \ref{app_prop:vincent}. $T_1$ represents terms do not contain $\theta$.

The next equality between \eqref{eq:thm pf d} and \eqref{eq:thm pf e} applies the Boltzmann Distribution in Assumption \ref{assump:Boltzmann_Dist}. 

The equivalence in \eqref{eq:thm pf f} holds because the constant factors $T_1$ and $\frac{\tau^4}{(kT)^2}$ do not affect the optimization for $\theta$. Besides, $\widetilde{\widetilde{GNN}}_{\theta}$ and $GNN_{\theta}$ are up to a constant factor, which is viewed as equivalence in \cite{SheheryarZaidi2022PretrainingVD}. Another understanding is that the two GNNs learn the same force field label but up to unit conversion.

% \begin{subequations}
% \begin{align}
%     \simeq &E_{p (x_f)}||\widetilde{GNN}_{\theta} (x_f) - \nabla _{x_f} \log p (x_f)||^2 \\
%     =&E_{p (x_f,x_m)}||\widetilde{GNN}_{\theta} (x_f) - \nabla _{x_f} \log p (x_f|x_m)||^2+T_1 \\
%     =&E_{p (x_f,x_m,x_{eq})}||\widetilde{GNN}_{\theta} (x_f) - \nabla _{x_f} \log p (x_f|x_m)||^2+T_1  \\
%     =&E_{p (x_f|x_m)p(x_m|x_{eq})p(x_{eq})}||\widetilde{GNN}_{\theta} (x_f) - \frac{x_m-x_f}{\tau^2})||^2+T_1  \\
%     \simeq &E_{p (x_f|x_m)p(x_m|x_{eq})p(x_{eq})}||GNN_{\theta} (x_f) - \frac{-kT}{\tau^2}(x_f-x_m))||^2+T_1.  
% \end{align}
% \end{subequations}
\end{proof}
One may think about denoising the whole hybrid noise rather than fractional denoising. The following result shows when the whole hybrid noise is not coordinate Gaussian noise, we can not establish the equivalence between denoising and force learning. 
    \begin{equation}\label{eq:app b3}
    \begin{aligned}
         \eqref{eq:thm pf d}&=\mathbb{E}_{p (x_f,x_{eq})}\tau^4||\widetilde{GNN}_{\theta} (x_f) - \nabla_{x_f} \log p (x_f|x_{eq})||^2+T_2 \\
        &\neq \mathbb{E}_{p (x_f,x_m,x_{eq})}||GNN_{\theta} (x_f) - (x_f-x_{eq})||^2+T_2
    \end{aligned}
    \end{equation}
   A concrete investigation is provided in section \ref{sec:neccessity}. 
\begin{lemma}[The equivalence between score matching and conditional score matching~\citep{PascalVincent2011ACB}]
\label{app_prop:vincent}
    The following two optimization objectives with respect to $\theta$ are equivalent, i.e. $J_1(\theta)\simeq J_2(\theta)$.
    \begin{equation}
        J_1(\theta)=\mathbb{E}_{p (\tilde{x})}||GNN_{\theta} (\tilde{x}) - \nabla _{\tilde{x}} \log p (\tilde{x})||^2 
    \end{equation}
     \begin{equation}
         J_2(\theta)= \mathbb{E}_{p (\tilde{x}|x)p(x)}||GNN_{\theta} (\tilde{x}) - \nabla _{\tilde{x}} \log p (\tilde{x}|x)||^2
     \end{equation}
\end{lemma}
\begin{proof}
    We first expand the square term and observe:
    \begin{equation}
    J_1(\theta)=\mathbb{E}_{p (\tilde{x})}[||GNN_{\theta} (\tilde{x})||^2] - 2\mathbb{E}_{p (\tilde{x})}[\left <GNN_{\theta} (\tilde{x}),\nabla _{\tilde{x}} \log p (\tilde{x})\right > ]+T_3
    \end{equation}
     \begin{equation}
         J_2(\theta)= \mathbb{E}_{p (\tilde{x}|x)p(x)}[||GNN_{\theta} (\tilde{x})||^2] - 2\mathbb{E}_{p (\tilde{x}|x)p(x)}[\left <GNN_{\theta} (\tilde{x}),\nabla _{\tilde{x}} \log p (\tilde{x}|x)\right > ]+T_4,
     \end{equation}
    where $T_3$, $T_4$ are constants independent of $\theta$. Therefore, it suffices to show that the middle terms on the right-hand side are equal.
    \begin{equation}
        \begin{aligned}
        &\mathbb{E}_{p (\tilde{x})}[\left <GNN_{\theta} (\tilde{x}),\nabla _{\tilde{x}} \log p (\tilde{x})\right > ]\\
        =& \int_{\tilde{x}} p (\tilde{x})  \left <GNN_{\theta} (\tilde{x}),\nabla _{\tilde{x}} \log p (\tilde{x})\right >  \mathrm{d}\tilde{x} \\
         =& \int_{\tilde{x}} p (\tilde{x})  \left <GNN_{\theta} (\tilde{x}),\frac{\nabla _{\tilde{x}}  p (\tilde{x})}{p (\tilde{x})}\right >  \mathrm{d}\tilde{x} \\
         =& \int_{\tilde{x}}  \left <GNN_{\theta} (\tilde{x}),\nabla _{\tilde{x}}  p (\tilde{x})\right >  \mathrm{d}\tilde{x} \\
         =& \int_{\tilde{x}}  \left<GNN_{\theta} (\tilde{x}),\nabla _{\tilde{x}} \left (\int_{x} p (\tilde{x}|x)p(x) \mathrm{d}x\right ) \right > \mathrm{d}\tilde{x} \\
         =& \int_{\tilde{x}}  \left <GNN_{\theta} (\tilde{x}), \int_{x} p(x) \nabla _{\tilde{x}} p (\tilde{x}|x) \mathrm{d}x \right >  \mathrm{d}\tilde{x} \\
         =& \int_{\tilde{x}}  \left <GNN_{\theta} (\tilde{x}), \int_{x} p (\tilde{x}|x) p(x) \nabla _{\tilde{x}}\log p (\tilde{x}|x)  \mathrm{d}x \right >  \mathrm{d}\tilde{x} \\
        =& \int_{\tilde{x}} \int_{x} p (\tilde{x}|x) p(x)  \left <GNN_{\theta} (\tilde{x}), \nabla _{\tilde{x}}\log p (\tilde{x}|x)  \right >   \mathrm{d}x \mathrm{d}\tilde{x} \\
        =& \mathbb{E}_{p (\tilde{x},x)} [ \left <GNN_{\theta} (\tilde{x}), \nabla _{\tilde{x}}\log p (\tilde{x}|x)  \right > ]\\
        \end{aligned}
    \end{equation}
    
\end{proof}

\subsection{Proofs for~\ref{tab:ff}}\label{section:appendix proof table}
Define a transformation function for a molecule $\mathcal{M}$ that maps from coordinates to its bonds, angles, and torsion angles:
\begin{align*}
     f_{VRN}^\mathcal{M}:\mathbb{R}^{3N}&\longrightarrow (\mathbb{R}_{\geq 0})^{m_1}\times([0,\pi])^{m_2}\times([0,2\pi))^{m_3}\times([0,2\pi))^{m}  \in \mathbb{R}^{m_1+m_2+m_3+m} \\
    x &\longmapsto  d_{VRN}=(r,\theta,\phi,\psi)
\end{align*}
and another transformation function that maps from coordinates to its rotatable torsion angles:
\begin{align*}
     f_{RN}^\mathcal{M}:\mathbb{R}^{3N}&\longrightarrow ([0,2\pi))^{m}   \in \mathbb{R}^{m}  \\
    x &\longmapsto  \psi
\end{align*}
 The functions are well-defined since the bonds, angles, and torsion angles are uniquely determined by the coordinates.
 % Also, the transformation functions are differentiable at regular conformations. 
 Then we present a theorem that establishes a relationship between the (vibration) rotation noise and the coordinate changes of a given conformation.
\begin{theorem}[Noise type transformation]\label{thm:Noise type transformation}
Let $x$ be an input conformation and let $\Delta d_{VRN}$ be vibration rotation noise (VRN) and $\Delta \psi$ be rotation noise (RN), respectively. $x_{m}$ is the resulting conformation obtained by adding the VRN or RN to $x$. Then, under certain conditions, the following holds:

For VRN: When $\Delta d_{VRN} \to 0$, the corresponding coordinate changes $\Delta x = x_m - x$ are asymptotically linear to the VRN, i.e., $\Delta x = C_{VRN}(x)\Delta d_{VRN} + o(\Delta d_{VRN})$, where o() refers to infinitesimal of higher order.

For RN: When $\Delta \psi \to 0$, the corresponding coordinate changes $\Delta x = x_m - x$ are asymptotically linear to the RN noise, i.e., $\Delta x = C_{RN}(x)\Delta \psi + o(\Delta \psi)$.
\end{theorem}
     Here we provide a general proof for VRN and RN. It is a direct application of the definition of the differential of a multivariable function. Later we will provide a constructive proof for T-rot noise to gain more intuitive insights in theorem~\ref{thm:Noise type transformation constructive}. Generally, let $f$ be the transformation function, $d=f(x)$, and let $M$ be the image space dimension of $f$, i.e. $M=\sum_{i=1}^3 m_i+m$ for $f=f_{VRN}^\mathcal{M}$ and $M=m$ for $f=f_{RN}^\mathcal{M}$. According to the detailed implementation of VRN and RN in section~\ref{app sec:algorithms}, we have $M\leq 3N$ always holds. The result to be proved is $\Delta x = C(x)\Delta d + o(\Delta d)$.
     
      The regularity conditions for Theorem~\ref{thm:Noise type transformation} are:\\
         1.  The transformation function is differentiable at input conformation $x$.\\
         2.  The function $A$ is differentiable at $x^{(1)}$, where $A$ and $x^{(1)}$ will be defined in the proof.\\
         3.  By rearranging the order of dimensions of $x$, we can obtain a partition such that $(J^{(1)}+J^{(2)}J_A)$ is full rank, where $J^{(1)}$,$J^{(2)}$ and $J_A$ will be defined in the proof.  

     We will now proceed to prove the theorem using these conditions. 
\begin{proof}
     According to the definition of differential, for a differentiable point $x$ of $f$, we have: 
     \begin{equation}
         \Delta d=f(x+\Delta x)-f(x) = J(x)\Delta x + \alpha(x;\Delta x).
     \end{equation}
     When $\Delta x\to 0, \alpha(x;\Delta x)=o(\Delta x)$.
     $J(x)$ is the differentiation of $f$ at point $x$. It can be written as a 
$M\times 3N$ Jacobi matrix: 
     \begin{equation}
        J(x)=\left(             
                \begin{array}{cccc}
                \frac{\partial f^{1}}{\partial x^1} &\cdots  & \frac{\partial f^{1}}{\partial x^{3N}} \\ 
                 \vdots  & \ddots & \vdots \\ 
                 \frac{\partial f^{M}}{\partial x^1} &\cdots  & \frac{\partial f^{M}}{\partial x^{3N}}
                \end{array}
        \right ).
\end{equation}
 Following the partition in the third regularity condition, $J\Delta x=[J^{(1)},J^{(2)}] [\Delta x^{(1)},\Delta x^{(2)}]^\top$, where $J^{(1)}$ is a $M\times M$ matrix, $J^{(2)}$ is a $M\times (3N-M)$ matrix, $\Delta x^{(1)}\in\mathbb{R}^{M},\Delta x^{(2)}\in \mathbb{R}^{3N-M}$.
 
Since the noise has $M$ degree of freedom, we can conclude that $x_m$ has $(3N-M)$ equations as constraints given $x$. Therefore, $x_m^{(2)}$ can be uniquely determined by $x_m^{(1)}$, allowing us to express $x_m^{(2)}$ as a function of $x_m^{(1)}$ as follows: $x_m^{(2)} = A(x_m^{(1)})$, where $A$ is a function dependent on the initial input $x$.
 As assumed in the second regularity condition, $A$ is differentiable at $x^{(1)}$, then when $\Delta x^{(1)}\to 0$,
 \begin{equation}
     x^{(2)}_m-x^{(2)}=\Delta x^{(2)}=J_A(x^{(1)})\Delta x^{(1)}+ o(\Delta x^{(1)}).
 \end{equation}  
 According to the third regularity condition, $B(x)\triangleq J^{(1)}(x)+J^{(2)}(x)J_A(x^{(1)})$ is a full rank $M\times M$ matrix, so it has an inverse matrix. 
  \begin{equation}
  \begin{aligned}
       \Delta d &= J^{(1)}\Delta x^{(1)}+J^{(2)}\Delta x^{(2)} + o(\Delta x)\\
     &=J^{(1)}\Delta x^{(1)}+J^{(2)}J_A\Delta x^{(1)}+ J^{(2)}o(\Delta x^{(1)})+ o(\Delta x)\\
     &=B \Delta x^{(1)}+ o(\Delta x)
  \end{aligned}
 \end{equation} 
\begin{equation}
  \begin{aligned}
     \Delta x^{(1)} &= B^{-1} \Delta d - B^{-1}  o(\Delta x)\\
     &= B^{-1} \Delta d+ o(\Delta x)\\
  \end{aligned}
 \end{equation} 
 \begin{equation}
  \begin{aligned}
      \Delta x &=\left(             
                \begin{array}{c}
                \Delta x^{(1)} \\ 
                 \Delta x^{(2)}
                \end{array}
        \right )
        =\left(             
                \begin{array}{c}
                I \\ 
                J_A
                \end{array}
        \right ) \Delta x^{(1)}
        = \left(             
                \begin{array}{c}
                I \\ 
                J_A
                \end{array}
        \right )\left(B^{-1} \Delta d+ o(\Delta x)\right )\\
        &= \left(             
                \begin{array}{c}
                I \\ 
                J_A
                \end{array}
        \right )B^{-1} \Delta d+ o(\Delta x) \triangleq C \Delta d+ o(C \Delta d)= C \Delta d+o(\Delta d),
  \end{aligned}
 \end{equation} 
 where $C(x)=\left(             
                \begin{array}{c}
                I \\ 
                J_A (x^{(1)})
                \end{array}
        \right )B^{-1}(x)$.
\end{proof}

Now we consider the conditional probability distributions, which indicate the forces the GNN learned by denoising. The notations are consistent with section~\ref{section:Hybrid noise design}. We propose two kinds of chemical-aware noise (CAN): $\Delta d\sim \mathcal{N}(d_{eq}, \Sigma)$. For VRN, $\Sigma=\Sigma(\sigma_r,\sigma_{\theta},\sigma_{\phi},\sigma_{\psi})=
    \begin{tiny}
        \left(             
                    \begin{array}{cccc}
                    \sigma_r^2  I_{m_1}  &\bm{0} & \bm{0} & \bm{0} \\ 
                     \bm{0} & \sigma_{\theta}^2  I_{m_2} & \bm{0} & \bm{0}\\
                     \bm{0} & \bm{0} &\sigma_{\phi}^2  I_{m_3} & \bm{0}\\
                     \bm{0} & \bm{0} & \bm{0} &\sigma_{\psi}^2  I_m
                    \end{array}
            \right )
    \end{tiny}$.
    For RN, $\Sigma=\Sigma(\sigma)=\sigma^2  I_m$. coordinate Gaussian noise (CGN) is $p(x_f|x_m)\sim \mathcal{N}(x_m,\tau^2  I_{3N})$.
\begin{theorem}\label{thm:gauss1}
    Under the same conditions in theorem \ref{thm:Noise type transformation}, when $\Delta d\to 0$, the asymptotic conditional probability distribution corresponding to CAN is $ p(x_m|x_{eq})\sim \mathcal{N}(x_{eq},  \Gamma_1(x) )$, where $\Gamma_1(x)=C(x)\Sigma C(x)^\top$ ,$C(x)$ is given by theorem \ref{thm:Noise type transformation}.
    The asymptotic conditional probability distribution corresponding to the entire hybrid noise is $p(x_f|x_{eq})\sim \mathcal{N}(x_{eq}, \Gamma_2(x))$, where $\Gamma_2(x)= \tau^2 I_{3N}+C(x)\Sigma C(x)^\top$.
\end{theorem}
\begin{proof}
    From theorem \ref{thm:Noise type transformation}, we obtain the asymptotic linearity $x_m=x_{eq}+C\Delta d+o(d)$, when $\Delta d\to 0$.
    Since a linear transformation of a Gaussian random variable is still a Gaussian random variable, $p(x_m|x_{eq})$ is asymptotically a Gaussian random variable. The asymptotic covariance is 
        $ cov(x_m)= cov(C\Delta d)
        = C cov(\Delta d)C^\top 
        = C \Sigma C^\top$.
    Therefore $p(x_m|x_{eq})\sim \mathcal{N}(x_{eq},C\Sigma C^\top )$.

    As for CGN, $x_f=x_m + \tau \epsilon$, where $\epsilon\sim \mathcal{N}(0,I_{3N})$. So $x_f= x_{eq}+C\Delta d + \tau \epsilon +o(d)$, where $\epsilon \upmodels \Delta d$ are two independent Gaussian random variables. Since the sum of independent Gaussian random variables is still a Gaussian random variable, $p(x_f|x_{eq})$ is a Gaussian random variable. 
     The asymptotic covariance is 
        $cov(x_f)=cov(\tau \epsilon +C\Delta d) = \tau^2 I + C\Sigma C^\top$.
\end{proof}
In the subsequent theorem, we present the noise type transformation specific to RN, wherein $C_{T}(x)$ is explicitly constructed. The proof facilitates an intuitive grasp of the dependency of the linear coefficient matrix on the input conformation. 
\begin{theorem}[Noise type transformation for rotation noise with a constructive proof]
\label{thm:Noise type transformation constructive}
    Consider adding dihedral angle noise $\Delta\psi$ on the input structure $x_{eq}$. The corresponding coordinate change $ \Delta x=x_m-x_{eq}$ is approximately linear with respect to the dihedral angle noise, when the scale of the dihedral angle noise is small.
    \begin{equation}\label{Noise type transformation}
      ||\Delta x-C\Delta \psi||_2^2\leq \sum_{j=1}^m D_j \mathcal{E}(\Delta \psi_j)
    \end{equation}
     where  $\Delta\psi\in [0,2\pi)^m$, $\Delta x\in \mathbb{R}^{3N}$, $C$ is a $3N\times m$ matrix that is dependent on the input conformation, $\{D_j, j=1\cdots m\}$ are constants dependent on the input conformation. 
     \begin{equation}
     \begin{aligned}
         &\mathcal{E}(\Delta\psi_j) = (\Delta\psi_j)^2-2\Delta\psi_j \sin \Delta\psi_j-2cos\Delta\psi_j+2 \\
     &= (\Delta\psi_j)^2-2\Delta\psi_j (\Delta\psi_j+O((\Delta\psi_j)^3))+2(\frac{(\Delta\psi_j)^2}{2}+O((\Delta\psi_j)^4)) \\
     &=O((\Delta\psi_j)^4)
     \end{aligned}
     \end{equation}
     Therefore, $\Delta x=C\Delta \psi+o(\Delta \psi) $, indicating the linear approximation error is an infinitesimal of higher order, when $\Delta \psi\to 0$. If we further assume the rotations are not trivial, i.e. rotation of the rotatable bonds causes the movement of some atom positions, the rank of $C$ is $m$. All the elements above the main diagonal in $C$ are zero.  
\end{theorem}
\begin{proof}
To analyze the coordinate change after altering the dihedral angles of all the rotatable bonds in a molecule, we first consider changing one dihedral angle. As a proof in elementary geometry, we define some notations: $ \overline{AA'}$ is the length of the line segment $AA'$, $\wideparen{AA'}$ is the length of the arc segment $AA'$, $\angle$ represents angle, $\triangle A'AA''$ represents a triangle formed by point A', A and A'',$\perp$ represents perpendicular.
\begin{lemma}\label{lm:distance_propto}
    If the change in one dihedral angle $\Delta \psi$ is small enough, the distance the associated atoms move is proportional to the amount of the change in that dihedral angle, and the proportional coefficient is dependent on the input conformation.%, i.e. $\overline{XX'}\propto \psi$. The associated atoms are atoms that there coordinates are changed when altering that dihedral angle.$ \overline{XX'}$ is the distance the atom $X$  move.
\end{lemma}
\begin{proof}
For example, in \ref{fig:aspirine_cone_fig}, we study the effect on the coordinates of atom $A$ of turning the dihedral angle $\psi_1$, i.e. $\angle{ A O_A A'}  =\Delta \psi_1$. If the scale of the dihedral angle noise $\Delta\psi_1$ is small, the distance can be approximated by arc length.
\begin{equation}\label{eq6}
    \overline{AA'}\approx  \wideparen{AA'} = (\overline{OA}\cos\angle{ O A O_A})\Delta \psi_1,
\end{equation}
Please note that $\overline{OA}$ and $\angle{ O A O_A}$ are both determined by the original structure and remain constant when changing the dihedral angle. Therefore, $\overline{AA'}\propto \Delta\psi_1$. 

The same can be proved for other associated atoms. For example for atom $B$, with $\angle {B O_B B'} =\Delta \psi_1$ and $\overline{BB'}\approx \wideparen{BB'} = (\overline{OB}\cos\angle{O B O_B})\Delta \psi_1$,  we can deduce $\overline{BB'}\propto \Delta\psi_1$.

In conclusion, the distance the associated atoms move is proportional to the amount of change in the dihedral angle.
\end{proof}
Then, we extend this conclusion from distance to coordinates. Note that in lemma \ref{lm:coordinate_propto}, the notation $x$ represents one coordinate of the 3D coordinates, relative to $y$ and $z$.
\begin{lemma}\label{lm:coordinate_propto}
    If the change in one dihedral angle $\Delta \psi$ is small enough so that the distance the associated atoms move is also small, then the coordinate changes of the associated atoms are also proportional to the amount of the change in that dihedral angle, and the proportional coefficient is dependent on the input conformation.
\end{lemma}
\begin{proof}
Denote the 3D coordinate of atom $A$ as $(x,y,z)$. When changing the dihedral angle, the atom lies on the circle with center $O_A$, radius $\overline{O_A A}$ and perpendicular to $\overline{O_A O}$, i.e. $(x,y,z)$ satisfies
\begin{numcases}{}
(x-x_{O_A})^{2}+(y-y_{O_A})^{2}+(z-z_{O_A})^{2}=\overline{O_A A}^2\label{sphere}\\
(x-x_{O_A})(x_O-x_{O_A})+(y-y_{O_A})(y_O-y_{O_A})+(z-z_{O_A})(z_O-z_{O_A})=0.\label{plane}
\end{numcases}
Considering a sufficiently small amount of coordinate movement, $(x,y,z)$ also satisfies the formula after differentiating the equation \eqref{sphere}\eqref{plane}.
\begin{numcases}{}\label{d_sphere}
2(x-x_{O_A})\Delta x+2(y-y_{O_A})\Delta y+2(z-z_{O_A})\Delta z=0 \\
\label{d_plane}(x_O-x_{O_A})\Delta x+(y_O-y_{O_A})\Delta y+(z_O-z_{O_A})\Delta z=0.
\end{numcases}
Since $OA\perp OO_A$, \eqref{d_sphere}\eqref{d_plane} are not linearly related. Then $\Delta x$, $\Delta y$, $\Delta z$ are in linear relationship with each other.
\begin{numcases}{}
\Delta y=C_x^y\Delta x \label{xy}\\
\Delta z=C_x^z\Delta x \label{xz},
\end{numcases}
where the constants $C_x^y=-\frac{(x-x_{O_A})(z_O-z_{O_A})-(x_O-x_{O_A})(z-z_{O_A})}{(y-y_{O_A})(z_O-z_{O_A})-(y_O-y_{O_A})(z-z_{O_A})}$, $C_x^z=\frac{(x-x_{O_A})(y_O-y_{O_A})-(x_O-x_{O_A})(y-y_{O_A})}{(y-y_{O_A})(z_O-z_{O_A})-(y_O-y_{O_A})(z-z_{O_A})}$. \\
With lemma \ref{lm:distance_propto} and the relationship between the coordinates and the distance, we obtain
\begin{equation}\label{xyz_psi}
    (\Delta x) ^2+(\Delta y) ^2+(\Delta z) ^2=\overline{AA'}^2 \propto \Delta\psi_1^2.
\end{equation}
Substituting \eqref{xy}\eqref{xz} into \eqref{xyz_psi}, we conclude 
\[\Delta x\propto \Delta y\propto \Delta z\propto \Delta\psi_1.\]
The same can be proved for other associated atoms. Therefore the coordinate changes are proportional to the amount of the change in the dihedral angle.
\end{proof}
\begin{figure}[thbp]
\vskip 0.2in
\begin{center}
\centerline{\includegraphics[width=\textwidth]{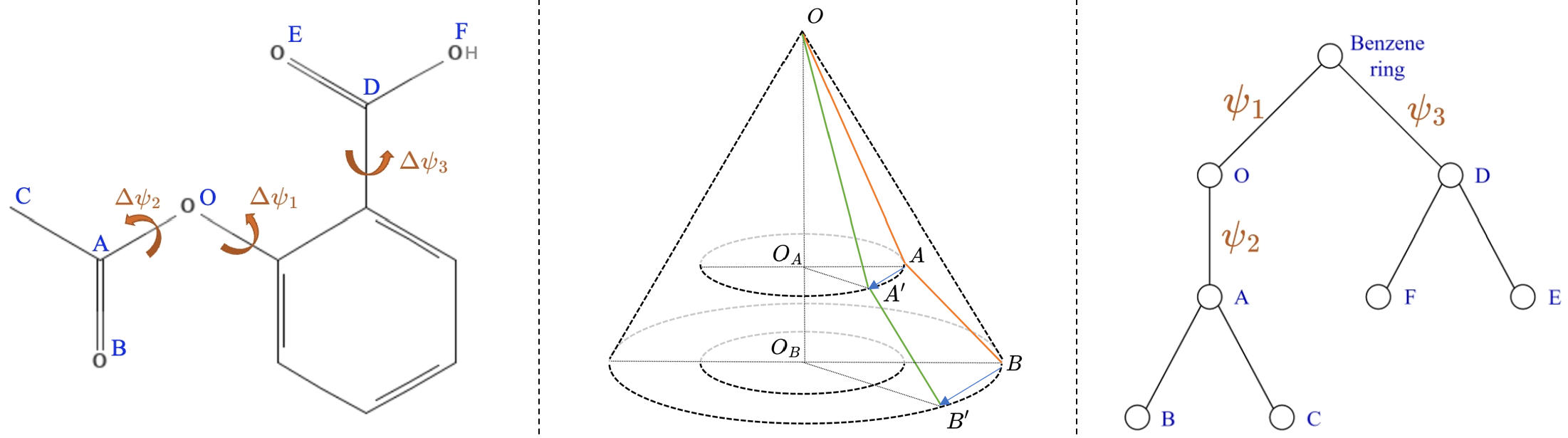}}
\caption{\textbf{\thefigure:} \textbf{Illustrations to aid the proof of Theorem \ref{thm:Noise type transformation constructive}.} Left: Three rotatable bonds in aspirin. Middle: When changing the dihedral angle $\psi_1$, the atoms move along a circular arc e.g. $A\rightarrow A'$ and $B \rightarrow B'$. Right: When considering the changing of all dihedral angles, we can define a breadth-first order to traverse all rotatable bonds in the tree structure of aspirin, e.g. $(\psi_1,\psi_3,\psi_2)$, and consider their effects on coordinate one by one and then add the effects together.}
\label{fig:aspirine_cone_fig}
\end{center}
\vskip -0.2in
\end{figure}
The proof in lemma \ref{lm:distance_propto} and lemma \ref{lm:coordinate_propto} requires ``$\Delta\psi_1$ is sufficiently small", indicating that the linear relationship between two types of noise is an approximation. Here we specify the approximation error.
\begin{lemma}\label{lm:linear approx error}
    The approximation error for atom A is given by $ \overline{O_A A}^2 \mathcal{E}(\Delta\psi_1)=  (\overline{OA}\cos\angle{ O A O_A})^2 \mathcal{E}(\Delta\psi_1)$, where $\mathcal{E}(\Delta\psi_1)= [\Delta\psi_1^2-2\Delta\psi_1 \sin \Delta\psi_1-2cos\Delta\psi_1+2]$ is the term only dependent on $\Delta\psi_1$ in the error. $\lim_{\Delta\psi_1\to 0} \mathcal{E}(\Delta\psi_1)=0$, indicating the approximation error is small when $\Delta\psi_1$ is small. $\overline{O_A A}=\overline{OA}\cos\angle{ O A O_A}$ is determined by the molecular structure. 
\end{lemma}
 \begin{proof}
    The approximation used in \eqref{eq6}, \eqref{d_sphere}, \eqref{d_plane} is summarized by approximating $\vec{AA'} $ by $\vec{AA''} $ in \ref{fig:error_approx}, where $\vec{AA'} $ is the coordinate change 
\begin{minipage}[b]{0.72\textwidth}
    % \begin{proof}
    % The approximation used in \eqref{eq6}, \eqref{d_sphere}, \eqref{d_plane} is summarized by approximating $\vec{AA'} $ by $\vec{AA''} $ in Figure \ref{fig:error_approx}, 
     after adding dihedral angle noise while $\vec{AA''} $ is the approximated coordinate change that is linear to the dihedral angle noise, $\overline{AA''}=\wideparen{AA'}$ (i.e. \eqref{eq6}) and $AA''$ is tangent to circle $O_A$ at point $A$ (i.e. \eqref{d_sphere}, \eqref{d_plane}). Therefore, $\overline{AA'}=2\overline{O_A A}sin(\frac{\Delta\psi_1}{2})$, $\angle{ A' A A''}=\frac{\Delta\psi_1}{2}$, $\overline{AA''}=\wideparen{AA'}=\overline{O_A A}\Delta\psi_1$. By using the law of cosines in $\triangle A' A A''$, we have the approximating error $||\vec{AA'} -\vec{AA''}||_2^2= ||\vec{A''A'}||_2^2=\overline{A'A''}^2=\overline{O_A A}^2[(2sin(\frac{\Delta\psi_1}{2}))^2+(\Delta\psi_1)^2-2(2sin(\frac{\Delta\psi_1}{2}))\Delta\psi_1 cos(\frac{\Delta\psi_1}{2})]=\overline{O_A A}^2[2(1-cos(\Delta\psi_1))+(\Delta\psi_1)^2-2\Delta\psi_1sin(\Delta\psi_1)]$. $\lim_{\Delta\psi_1\to 0} [2(1-cos(\Delta\psi_1))+(\Delta\psi_1)^2-2\Delta\psi_1sin(\Delta\psi_1)]=0$.
\end{minipage}
\hfill
\begin{minipage}[b]{0.23\textwidth}
\centering
  \begin{figure}[H]
  \centering
  \vskip -0.2in
    \includegraphics[width=1\textwidth]{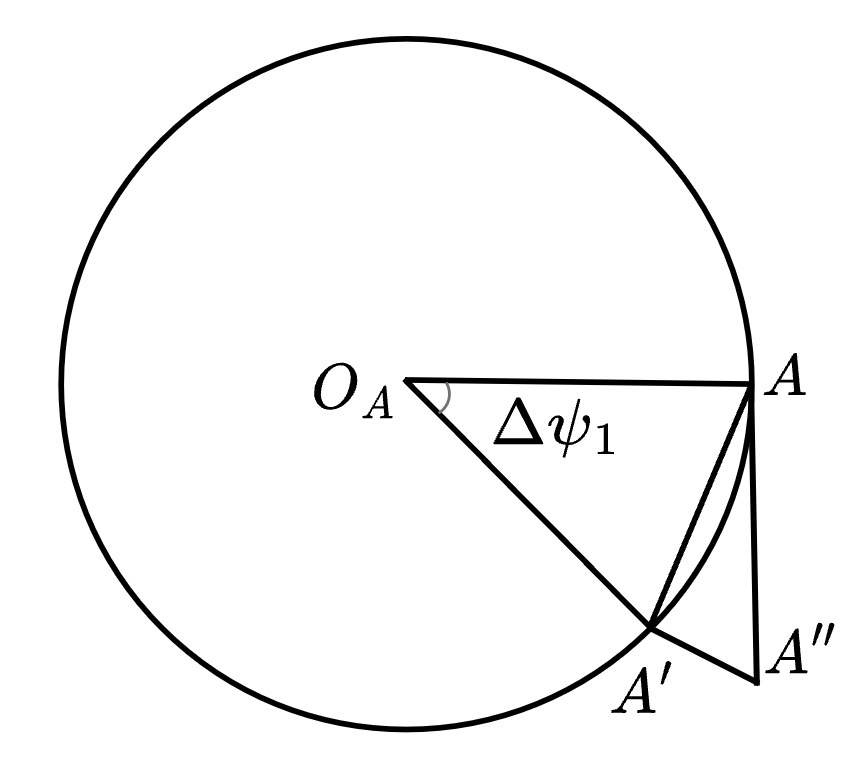}
    \caption{\textbf{\thefigure:} Illustrations to aid the proof of Lemma \ref{lm:linear approx error}}
    \label{fig:error_approx}
  \end{figure}
\end{minipage}
% \begin{wrapfigure}{r}{2cm}
% \vskip -0.2in
% \begin{center}
% % \centerline{\includegraphics[width=4cm]{error_approx.png}}
% % \caption{}
% % \label{fig:error_approx}
% \end{center}
% \vskip -0.2in
% \end{wrapfigure}
% \end{proof}

    \end{proof}

Now we consider changing all the dihedral angles of the rotatable bonds. Making use of lemma~\ref{lm:coordinate_propto}, the overall changes in coordinates are the sum of the coordinate change caused by each dihedral angle. Denote the linear coefficients as C, we can obtain $\lim_{\Delta \psi\to 0}||\Delta x-C\Delta \psi||=0$, and C is dependent on the input conformation.

In fact, by specifying the form of $\Delta \psi$ and $\Delta x$ in \eqref{Noise type transformation}, we can further prove all the elements above the main diagonal in $C$ are zero and full rank. 
On the one hand, a molecule forms a tree structure, if rings and other atoms in the molecule are regarded as nodes, and the bonds are regarded as edges. We can define a breadth-first order to traverse all rotatable bonds. We arrange the dihedral angles in the vector $\Delta \psi$ in this order.
On the other hand, when rotating each bond sequentially, we consider the effects of the dihedral angles on its child nodes and keep the parent nodes still. Define one of the nearest atoms affected by the dihedral angle as the key atom of the dihedral angle. We put the coordinates of the key atoms in the first few components in the vector $\Delta x$ in \eqref{Noise type transformation} and arrange them correspondingly in the order of the dihedral angles they belong to. 
All the elements above the main diagonal in $C$ are zero because in the breadth-first order, later dihedral angles do not affect key atoms of the earlier dihedral angles. Also, we assume the rotations are not trivial, so every rotation of the rotatable bonds has a key atom, and thus the diagonal blocks ($3 \times 1$ submatrices) of $C$ are not $0$ submatrices ($3 \times 1$ submatrix whose elements are all zeros). Coupled with the condition that the elements above the main diagonal are zero, we derive the conclusion that the matrix $C$ is full rank. 

Take aspirin as an example, as is shown in~\ref{fig:aspirine_cone_fig}. Without loss of generality, we discuss three rotatable bonds labeled in the figure. Represent aspirin as a tree (the representation is not unique), and traverse all rotatable bonds in breadth-first order $(\psi_1,\psi_3,\psi_2)$. Their corresponding key atoms are $(A, F, B)$.
For simplicity, denote $\Delta_k\triangleq(\Delta_{x_k},\Delta_{y_k},\Delta_{z_k})^\top$ as coordinate changes of atom $k$, $k\in\{A,B,C,E,F\}$, and denote $\Delta_{Other}$ as coordinate changes of the atoms whose coordinates remain unchanged after altering the rotating the rotatable bonds. Then the approximate relation between atomic coordinate changes and dihedral angle changes is given by
\begin{equation}
    \left( \begin{array}{c}
\Delta_A\\
\Delta_F\\
\Delta_B\\
\Delta_C\\
\Delta_E\\
\Delta_{Other}
\end{array} 
\right ) \approx 
    \left( \begin{array}{ccc}
c_{11} & 0 & 0\\
0 & c_{22} & 0\\
c_{31} & 0 & c_{33} \\
c_{41} & 0 & c_{43}\\
0 & c_{52} & 0 \\
0 & 0 & 0 
\end{array} 
\right )
    \left( \begin{array}{c}
\Delta\psi_1\\
\Delta\psi_3\\
\Delta\psi_2
\end{array} 
\right ),
\end{equation}
where $c_{ij}, i,j=1,2\cdots$ represents $3\times 1 $ submatrices and $0$ in the matrix represents $3\times 1 $ submatrix whose elements are all zeros.

The approximation error $||\Delta x-C\Delta \psi||_2^2=\sum_i||\Delta_{{atom}_i}-\sum_j c_{ij}\Delta \psi_j||_2^2 =\sum_i||\sum_j\Delta_{{atom}_i,\psi_j}-\sum_j c_{ij}\Delta \psi_j||_2^2 \leq \sum_{i,j}||\Delta_{{atom}_i,\psi_j}-c_{ij}\Delta \psi_j||_2^2 = \sum_{i,j} d_{{atom}_i,\psi_j}^2 \mathcal{E}(\Delta \psi_j)= \sum_{j} ( \sum_{i}d_{{atom}_i,\psi_j}^2 )\mathcal{E}(\Delta \psi_j) $, where $\Delta_{{atom}_i}$ is the total coordinate change for ${atom}_i$, $\Delta_{{atom}_i,\psi_j}$ is the coordinate change caused by $\Delta \psi_j$ for ${atom}_i$ (It varies with the order of changing the dihedral angles, but the sum $\Delta_{{atom}_i}=\sum_{j}\Delta_{{atom}_i,\psi_j}$ is unique defined.), $d_{{atom}_i,\psi_j}$ is the radius of the circular motion of atom $i$ when the dihedral angle of $\psi_j$ changes, e.g. $\overline{O_A A}$ in lemma \ref{lm:linear approx error}.
\end{proof}



\section{Supplementary experiments}\label{sec:supple exp}
\added{In this section, we augment our experiments to comprehensively validate our motivations, theories, method designs, and model robustness. To verify our motivation, we confirm in Section~\ref{section:app learn forcefield} that learning the force field serves as an effective molecular pre-training target. Certain downstream results suggest that Frad has reached the ceiling of the force learning pre-training approach. 
For the validation of our theories and method designs, we confirm the theoretical analysis in Section~\ref{sec:neccessity} within Section~\ref{sec:app can denoise}. Subsequently, in Section~\ref{sec:replace can}, we verify the flexibility of CAN design and propose that our CAN design may have already achieved optimal performance. 
Regarding the design of the downstream fine-tuning method, we validate the effectiveness of the Frad noisy nodes design in Section~\ref{section:exp NN}. Lastly, in terms of model robustness, we demonstrate in Section~\ref{sec:app Architecture robust} that Frad exhibits effectiveness across different model architectures.}

\subsection{Learning force field help the downstream tasks}\label{section:app learn forcefield}
In this section, we conduct the force pre-training experiment to assess the potential enhancement of downstream performance through force learning. As obtaining the true force labels can be time-consuming, we randomly select 10,000 molecules with fewer than 30 atoms from the PCQM4Mv2 dataset and calculate their precise force labels using DFT. We then pre-train the model by predicting these force labels, followed by fine-tuning the model on three sub-tasks from the QM9 and MD17 datasets. As shown in \ref{table:learn forcefield}, pre-training by learning \replaced{DFT}{the} forces notably improves the performance of downstream tasks, suggesting the motivation of denoising pre-training to learn better forces is reasonable. 
\added{We also compare the force learning pre-training with Frad and Coord pre-trained on the same subset of data. Since Frad learns a more accurate force field than Coord as validated in section~\ref{sec:force accuracy}, the results indicate that the more accurate the pre-training force field, the better the downstream performance. Frad shows significant improvement over Coord and achieves comparable results to pre-training with DFT forces on QM9, suggesting that Frad has reached the ceiling of pre-training by force learning. However, results on MD17 indicate that there is still room for improvement. }

\begin{table}[h!]
\centering
% \vspace{-10pt}
% \setlength{\tabcolsep}{3pt}
    \caption{\textbf{\thetable:} The performance (MAE) comparison between pre-training \replaced{methods}{with learning force field and training from scratch} on 3 sub-tasks from QM9 and MD17 datasets. The top results are in bold. }
    \label{table:learn forcefield}
    % \vskip 0.15in
    % \begin{center}
    % \begin{footnotesize}
    % \begin{sc}
    \begin{tabular}{lrrr}
    \toprule
    	 & QM9:\makecell[c]{$\epsilon_{HOMO}$ \\(meV)}		& QM9:\makecell[c]{$\epsilon_{LUMO}$\\ (meV)}		& MD17:Aspirin (Force)
     \\
    \midrule
training from scratch  & 19.2& 20.9&0.2530   \\ 
\added{Coord}  & 18.1& 20.7&0.2497   \\ 
\added{Frad}  & 17.3& 19.6&0.2476   \\ 
 pre-training \replaced{by learning DFT forces}{with learning force field} &\textbf{17.1}&\textbf{19.6}&\textbf{0.2361}  \\ 
    \bottomrule
    \end{tabular}
    % \end{sc}
    % \end{footnotesize}
    % \end{center}
    % \vskip -0.1in
    % \vspace{-10pt}
\end{table}

\subsection{\added{Comparing Frad and CAN denoising}}\label{sec:app can denoise}
\added{Our theoretical analysis in \ref{sec:neccessity} and \ref{tab:ff} shows that denoising just Chemical Aware Noise (CAN) does not preserve the force learning interpretation, which may compromise the performance of the model. To validate this conjecture, we conduct experiments to evaluate the performance of denoising just CAN. Specifically, we test two types of CAN: RN and VRN. The results are shown in \ref{apptable:denoisCAN}. The results indicate that pre-training with only VRN or only RN has no significant effect. The downstream results are similar to those obtained from training from scratch, with some cases even showing negative transfer. This suggests that direct denoising of CAN alone does not capture useful molecular knowledge and fractional denoising is necessary.}

\begin{table}[h!]
\centering
\caption{\added{\textbf{\thetable:} Test performance (MAE) of different pre-training on tasks in QM9 dataset. The best results are in bold and the negative transfer cases are underlined.}}
\label{apptable:denoisCAN}
    \vskip 0.15in
\begin{tabular}{lccccccccc}
\toprule
{Model} & {HOMO} & {LUMO} & {Gap} & {$U$} & {$U_0$} & {$H$} & {$G$} & {Average} \\
\midrule
Train from scratch & 20.3 & 18.6 & 36.1 & 6.4 & 6.2 & 6.2 & 7.6 & 14.5 \\
Denoise VRN        & 20.0 & 16.6 & 34.2 & \underline{8.1} & \underline{8.1} & \underline{7.9} & \underline{8.4} & 14.8 \\
Denoise RN         & \underline{20.5} & 16.3 & 33.5 & \underline{7.5} & \underline{7.3} & \underline{7.7} & \underline{7.9} & 14.4 \\
Frad(VRN)          & 17.9 & 13.8 & \textbf{27.7} & 5.4 & \textbf{5.4} & 6.0 & \textbf{6.0} & 11.7 \\
Frad(RN)           & \textbf{15.3} & \textbf{13.7} & 27.8 & \textbf{5.3} & 5.6 & \textbf{5.6} & 6.2 & \textbf{11.4} \\
\bottomrule
\end{tabular}
\end{table}

 \subsection{\added{Replacing CAN with Conformation Generation}}\label{sec:replace can}

\added{In our theory, CAN can be flexibly constructed, and its design goal is to make $p(x_{fin})$ more consistent with the molecular conformation distribution. In fact, a more accurate method is to use computational chemistry tools to sample molecular conformations. Theoretically, employing computational chemistry tools instead of CAN should yield a more accurate force field, potentially enhancing the effect of pre-training. This comparison can also help judge how much room for improvement in CAN designing under the Frad framework. }

\added{Therefore, we conduct a comparative experiment between denoising using CAN and conformation generation. To keep data generation time manageable, we select a random subset of the pre-training data comprising 100,000 molecules and use the RDKit toolkit to generate conformations via the Distance Geometry method, followed by optimization with the Merck molecular force field (MMFF). Since Frad is pre-trained for 8 epochs, with Chemical Aware Noise (CAN) added to each molecule once per epoch, for fairness, we generate 8 conformations for each molecule using RDKit. During training, one conformation is used per epoch, with a total of 8 epochs trained.}

\added{The model that uses generated low-energy conformations and performs coordinate denoising is referred to as Frad(RDKit+Coord). For comparison, we also pre-trained Frad on the same subset of data, denoted as Frad(CAN+Coord). Their performance comparison is shown in Table~\ref{apptab:RDKit+Coord}.}

\begin{table}[htbp]
\centering
\caption{\added{\textbf{\thetable}: Test performance (MAE) comparison of Frad and coordinate denoising with generated low-energy conformations. The best results are in bold.}}
\label{apptab:RDKit+Coord}
\begin{tabular}{l cccc}
\toprule
{Model} & {Homo(MeV)} & {Lumo(MeV)} & {Gap(MeV)} & {Average } \\
\midrule
Train from scratch & 20.3 & 18.6 & 36.1 & 25.0 \\
Frad(RDKit+Coord) & \textbf{18.5} & 17.1 &\textbf{ 31.6} & 22.4 \\
Frad(CAN+Coord) & \textbf{18.5} & \textbf{15.8} & 31.9 & \textbf{22.1} \\
\bottomrule
\end{tabular}
\end{table}

\added{Firstly, Frad(RDKit+Coord) surpasses training from scratch, indicating that replacing CAN with conformation generation is also effective. This aligns with our theory that the distribution of CAN can be flexibly designed to fit the molecular distribution. Secondly, both methods of fractional denoising perform comparably, with Frad(CAN+Coord) being slightly better on average. This suggests that even though the molecules obtained through Chemical Aware Noise (CAN) are not in their exact lowest energy conformations, they are sufficient for denoising pre-training to extract molecular knowledge. This also shows that our design of CAN has maximized the potential of the Frad framework. }

\added{Crucially, the process of generating low-energy conformations with computational chemistry tools is exceedingly time-consuming. We offer a speed comparison in \ref{apptab:time}. The slow speed of conformation generation makes large-scale pre-training impractical, thus highlighting the high efficiency of our method.}

\begin{table}[htbp]
\centering
\caption{\added{\textbf{\thetable}: Time comparison of generating low-energy conformations using RDKit and CAN.}}\label{apptab:time}
\begin{tabular}{p{4cm} p{3cm} p{1cm}}
\toprule
 \qquad  Sample Number  &  RDKit  &  CAN \\
\midrule
\qquad  100 & 2.77s & 0.86s \\
\qquad  1000 & 26.31s & 0.09s \\
\bottomrule
\end{tabular}
\end{table}

\subsection{Effectiveness of Frad noisy nodes}\label{section:exp NN}
Noisy Nodes \cite{godwin2021simple} is a method for improving downstream task performance by denoising during the fine-tuning phase. It involves corrupting the input structure with coordinate noise and training the model to predict the downstream properties and the noise from the same noisy structure. We have included the pseudocode for Noisy Nodes in Algorithm \ref{alg:nn}. However, we found that it cannot converge on force prediction tasks in the MD17 dataset with Noisy Nodes.

To address this issue, we proposed a series of modifications to help us understand why the traditional Noisy Nodes failed and improve it. 
We use the same model pre-trained by Frad (RN, $\sigma=2$, $\tau=0.04$) and fine-tune it on the Aspirin task in MD17 with distinct Noisy Nodes settings. The results are presented in \ref{table:ablation2}. 

Firstly, in the case of traditional Noisy Nodes (Setting 2), it fails to converge. 
Next, in Setting 3, we change the noise type into rotation noise and succeed in converging. We conjecture that this is because the task in MD17 is sensitive to the input conformation, whereas Noisy Nodes corrupt the input conformation by Coordinate noise leading to an erroneous mapping between inputs and property labels. This is validated by Setting 3, where the rotation noise has less perturbation on the forces, allowing the convergence.
Additionally, decoupling the input of denoising and downstream tasks (Setting 4) ensures an unperturbed input for downstream tasks, and fundamentally corrects the mapping, making it work effectively. 
Finally, further substitute the denoising task to be Frad (Setting 5) can further promotes the performance, indicating that learning a more accurate force field contributes to downstream tasks.
\begin{table}[h!]
\centering
% \vspace{-10pt}
    \caption{\textbf{\thetable:} Performance (MAE, lower is better) of different fine-tuning techniques on Aspirin task in MD17. NN denotes Noisy Nodes. DEC stands for decoupling the input of denoising and downstream tasks. The best results are in bold.  }
    \label{table:ablation2}
    % \vskip 0.15in
    % \begin{center}
    % \begin{footnotesize}
    % \begin{sc}
    \begin{tabular}{@{}p{2cm}p{5cm}p{2.5cm}@{}}
    \toprule
   Index & Settings & Aspirin (Force)\\
        \midrule       
        1  &   w/o NN 	&   0.2141 \\
       2  &  NN($\tau=0.005$)     & do not converge \\ 
      3  &   NN($\sigma=0.2$)   &	0.2096 \\
      4  &  NN($\tau=0.005$, DEC)     &  0.2107 \\ 
      % 5  &   \makecell[c]{NN($\sigma=0.2$, $\tau=0.005$, ALT) }	 &	0.2116\\
      5  &   \makecell[l]{NN($\sigma=20$, $\tau=0.005$, DEC) } 	 &	 \textbf{0.2087} \\
   \bottomrule
    \end{tabular}
    % \end{sc}
    % \end{footnotesize}
    % \end{center}
    % \vskip -0.1in
    % \vspace{-10pt}
\end{table}
% \vspace{-2pt}

\subsection{\added{Frad with other Network Architecture}}\label{sec:app Architecture robust}
\added{ Many equivariant molecular networks can be used in Frad.
To evaluate the robustness of our results to the architecture selection, we try our method on other commonly used equivariant networks, specifically EGNN and PaiNN. The results are shown in \ref{app table:egnn} and \ref{app table:painn}. The results show that Frad remains effective across different model structures, consistently and significantly outperforming training from scratch. Furthermore, our method exceeds the denoising pre-training baseline SE(3)-DDM, which also uses PaiNN as the network backbone. This demonstrates that the effectiveness of our method remains robust across different network architectures.}

\begin{table}[h!]
\centering
\caption{\added{\textbf{\thetable}: Test performance (MAE) of models with backbone EGNN on tasks in QM9 dataset. The best results are in bold.}}
\label{app table:egnn}
   \vskip 0.15in
\begin{tabular}{lcccc}
\toprule
{Model} & {Homo(MeV)} & {Lumo(MeV)} & {Gap(MeV)} & {Average } \\
\midrule
EGNN(Train from scratch) & 29.1 & 24.3 & 47.8 & 33.7 \\
Frad(EGNN)               & \textbf{22.1} & \textbf{21.1} & \textbf{36.8} & \textbf{26.7} \\
\bottomrule
\end{tabular}

\end{table}

\begin{table}[h!]
\centering
\caption{\added{\textbf{\thetable}: Test performance (MAE) of models with backbone PaiNN on tasks in QM9 dataset. The best results are in bold.}}
\label{app table:painn}
   \vskip 0.15in
\begin{tabular}{lcccc}
\toprule
{Model} & {Homo(MeV)} & {Lumo(MeV)} & {Gap(MeV)} & {Average } \\
\midrule
PaiNN(Train from scratch) & 26 & 22.5 & 45 & 31.2 \\
Frad(PaiNN)               & \textbf{21} & \textbf{18.1} & \textbf{37} & \textbf{25.4} \\
SE(3)-DDM(PaiNN)          & 23.48 & 19.42 & 40.22 & 27.7 \\
\bottomrule
\end{tabular}
\end{table}

\section{Experimental settings}\label{app: setting}
\subsection{Pre-training and fine-tuning algorithms}\label{app sec:algorithms}
We propose a physics-informed pre-training framework that encompasses a fractional denoising (Frad) pre-training method as well as fine-tuning methods that are compatible with denoising pre-training. 

In Frad pre-training, it involves adding a customizable chemical-aware noise(CAN), a coordinate Gaussian noise (CGN), and denoising the CGN noise. Frad pre-training is showcased by pseudocode Algorithm \ref{alg:frad}. 

\begin{algorithm}[h]
\caption{Fractional denoising pre-training}\label{alg:frad}
\begin{algorithmic}[1]
\Require
\Statex CAN-type: type of chemical-aware noise
\Statex$\sigma$, $\sigma_r$, $\sigma_\theta$, $\sigma_\phi$, $\sigma_{\psi}$: Scale of chemical-aware noise (CAN)
\Statex$\tau$: Scale of coordinate Gaussian noise (CGN)
\Statex${\rm Encoder}$: Encoder model
\Statex ${\rm NoiseHead}$: Network module for prediction of node-level noise
\Statex$X$: Pre-training dataset of molecular conformations
\Statex$T$: Training steps
% \FOR{i = 1,...,T}
\While{$T \neq 0$}
    \State $x_{eq}$ = dataloader($X$)  \Comment{random sample a molecular conformation from $X$}
    \If{CAN-type is RN }  
    \Comment{Adding CAN}
    \State Add Gaussian noise on the torsion angles of rotatable bonds in the molecule: $\psi_{med}=\psi_{eq}+\Delta\psi$, where $\Delta\psi \sim \mathcal{N}(0, {\sigma}^2I_{m})$ and obtain the conformation $x_{med}$.
    \Else[CAN-type is VRN] 
    \State Add Gaussian noise on the bond lengths, bond angles, and torsion angles: $r_{med}=r_{eq}+\Delta r$, $\theta_{med}=\theta_{eq}+\Delta\theta$, $\phi_{med}=\phi_{eq}+\Delta\phi$, $\psi_{med}=\psi_{eq}+\Delta\psi$ where $\Delta r \sim \mathcal{N}(0, {\sigma_r}^2I_{m_1})$, $\Delta\theta \sim \mathcal{N}(0, {\sigma_\theta}^2I_{m_2})$, $\Delta\phi \sim \mathcal{N}(0, {\sigma_\phi}^2I_{m_3})$, $\Delta\psi \sim \mathcal{N}(0, {\sigma_\psi}^2I_{m})$ and obtain the conformation $x_{med}$.
    \EndIf
    \State  $x_f = x_{med} + \Delta{x}$ ,  where $\Delta{x} \sim \mathcal{N}(0, {\tau}^2I_{3N})$
    \Comment{Adding CGN}
    \State$\Delta{x}^{pred} = {\rm NoiseHead}({\rm Encoder}(x_{fin}))$
    \State Loss = $||\Delta{x}^{pred} - \Delta{x}||_{2}^{2}$
    \State Optimise(Loss) \Comment{Update model parameters}
    \State $T = T - 1$
\EndWhile
% \ENDFOR
\end{algorithmic}
\end{algorithm}

In fine-tuning, we utilize Noisy Nodes techniques to further improve the performance. We apply traditional Noisy Nodes \cite{godwin2021simple} for tasks that is not very sensitive to conformations such as QM9. The traditional Noisy Nodes incorporates an auxiliary loss for coordinate denoising in addition to the original property prediction objective, as demonstrated in Algorithm \ref{alg:nn}. 

\begin{algorithm}[h]
\caption{Noisy Nodes algorithm\citep{godwin2021simple}}\label{alg:nn}
\begin{algorithmic}[1]
\Require
\Statex$\tau$: Scale of coordinate Gaussian noise (CGN)
\Statex${\rm Encoder}$: Encoder model
\Statex ${\rm NoiseHead}$: Network module for prediction of node-level noise
\Statex ${\rm PropHead}$: Network module for prediction of graph-level label
\Statex$X$: Training dataset containing molecular conformations and their property labels
\Statex$T$: Training steps
\Statex$\lambda_{p}$: Loss weight of property prediction loss
\Statex$\lambda_{n}$: Loss weight of Noisy Nodes loss
\While{$T \neq 0$}
    \State $x_{eq}, y_{eq}$ = dataloader($X$)  \Comment{random sample a pair of conformation and label from $X$}
    \State  $x_{fin} = x_{eq} + \Delta{x}$ , where $\Delta{x} \sim \mathcal{N}(0, {\tau}^2I_{3N})$ \Comment{Adding CGN}
    \State $y_{eq}^{pred}={\rm PropHead}({\rm Encoder}(x_{fin}))$
    \State $\Delta{x}^{pred}={\rm NoiseHead}({\rm Encoder}(x_{fin}))$
    \State Loss = $\lambda_{p}$PropertyPredictionLoss$(y_{eq}^{pred}, y_{eq})$+$\lambda_{n}||\Delta{x}^{pred} - \Delta{x}||_{2}^{2}$
    \State Optimise(Loss)
    \State $T = T - 1$
\EndWhile
% \ENDFOR
\end{algorithmic}
\end{algorithm}

As for tasks that are very sensitive to conformations such as force prediction tasks, traditional Noisy Nodes fail to converge. This is because traditional Noisy Nodes have to corrupt the input conformation leading to an erroneous mapping between inputs and property labels. Therefore, we utilize our proposed ``Frad Noisy Nodes" on the force prediction tasks such as MD17. Specifically, it firstly decouples the denoising task and the downstream task, and secondly substitutes Frad for the coordinate denoising. The algorithm is shown in Algorithm \ref{alg:frad_md} with distinctions from traditional Noisy Nodes highlighted in the color blue. The ablation study in section~\ref{section:exp NN} verifies the effectiveness of Frad Noisy Nodes. 

\begin{algorithm}[htbp]
\caption{Frad Noisy Nodes algorithm}\label{alg:frad_md}
\begin{algorithmic}[1]
\Require
\Statex CAN-type: type of chemical-aware noise
\Statex$\sigma$, $\sigma_r$, $\sigma_\theta$, $\sigma_\phi$, $\sigma_{\psi}$: Scale of chemical-aware noise (CAN)
\Statex$\tau$: Scale of coordinate Gaussian noise (CGN)
\Statex${\rm Encoder}$: Encoder model
\Statex ${\rm NoiseHead}$: Network module for prediction of node-level noise
\Statex ${\rm PropHead}$: Network module for prediction of graph-level label
\Statex$X$: Training dataset
\Statex$T$: Training steps
\Statex$\lambda_{p}$: Loss weight of property prediction loss
\Statex$\lambda_{n}$: Loss weight of Noisy Nodes loss
\While{$T \neq 0$}
    \State $x_{eq}, y_{eq}$ = dataloader($X$)  \Comment{random sample a pair of conformation and label from $X$}
    \State \textcolor{blue}{Add chemical-aware noise on $x_{eq}$ with parameter CAN-type and its corresponding scale, and obtain the intermediate conformation $x_{med}$.}
    \State  $x_{fin} = \textcolor{blue}{x_{med}} + \Delta{x}$ , where $\Delta{x} \sim \mathcal{N}(0, {\tau}^2I_{3N})$ \Comment{Adding CGN}
    \State $y^{pred} = {\rm PropHead}({\rm Encoder}(\textcolor{blue}{x_{eq}}))$
    \State$\Delta{x}^{pred} = {\rm NoiseHead}({{\rm Encoder}(x_{fin})})$
    \State Loss = $\lambda_{p}$PropertyPredictionLoss$(y^{pred}, y)$+$\lambda_{n}||\Delta{x}^{pred} - \Delta{x}||_{2}^{2}$
    \State Optimise(Loss)
    \State $T = T - 1$
    \EndWhile
% \ENDFOR
\end{algorithmic}
\end{algorithm}

The noise application including searching all the rotatable single bonds, perturbing the torsion angles, bond angles, bond lengths, and atomic coordinates in the molecule can be efficiently completed using RDKit, which is a fast cheminformatics tool~\cite{landrum2013rdkit,riniker2015better}. When searching rotatable bonds, the hydrogen atoms are taken into account, which can further improve the performance.
To maintain the independence of the noise, we make special treatments for distinct scenarios. Firstly, when an atom is connected to more than two atoms, we randomly select one edge and exclusively add noise to the angles involving this selected edge. Secondly, we do not add noise to the bonds, angles, and torsion angles that are inside the ring. As a consequence, the noise is independent and the degree of freedom of CAN does not exceed the number of atomic coordinates.

\subsection{Hyperparameter settings}\label{app sec:Hyperparameters}
% TODO: MD22 w/o NN

\begin{table}[h!]
\centering
    \caption{\textbf{\thetable:} Hyperparameters for Frad pre-training. }
    \label{table:app setting pretrain}
    % \vskip 0.15in
    % \begin{center}
    % \begin{footnotesize}
    % \begin{sc}
    \begin{tabular}{lr}
    \toprule
    Parameter & Value or description\\
     \midrule  
   Train Dataset & PCQM4MV2	\\
   Batch size & 	70	\\
    \midrule 
Optimizer  & 	AdamW	\\
% Weight Decay & 	0	\\
Warm up steps & 	10000	\\
Max Learning rate & 	0.0004	\\
Learning rate decay policy	 & Cosine\\
	Learning rate factor & 	0.8\\
	Cosine cycle length	 & 400000\\
 \midrule 
% Network structure	& \makecell[c]{Keep aligned with downstream settings \\respectively on QM9 and MD17}\\
%  \midrule 
% Nosiy node denoise weight  & 	1	
std. of torsion angles of rotatable bonds for CAN(RN) & 	2	\\
std. of bond lengths for CAN(VRN) & 	0.058	\\
std. of bond angles for CAN(VRN) & 	0.129	\\
std. of torsion angles of non-rotatable bonds for CAN(VRN) & 	0.18	\\
std. of torsion angles of rotatable bonds for CAN(VRN) & 	1	\\
std. of CGN  & 	0.04	\\
   \bottomrule
    \end{tabular}
    % \end{sc}
    % % \end{footnotesize}
    % % \end{center}
    % \vskip -0.1in
\end{table}
Hyperparameters for pre-training are listed in \ref{table:app setting pretrain}. 
Details about Learning rate decay policy can be referred in \href{https://hasty.ai/docs/mp-wiki/scheduler/reducelronplateau#strong-reducelronplateau-explained-strong}{https://hasty.ai/docs/mp-wiki/scheduler/reducelronplateau\#strong-reducelronplateau-explained-strong}.
\begin{table}[h!]
\centering
    \caption{\textbf{\thetable:} Hyperparameters for fine-tuning on MD17. }
    \label{table:app setting md17}
    % \vskip 0.15in
    % % \begin{center}
    % \begin{footnotesize}
    % \begin{sc}
    \begin{tabular}{lr}
    \toprule
    Parameter & Value or description\\
     \midrule  
  Train/Val/Test Splitting* &	9500/500/remaining data (950/50/remaining data)	\\
  Batch size* &	80 (8)	\\
  \midrule  
Optimizer&	AdamW	\\
% Weight Decay&	0	\\
Warm up steps	&1000	\\
Max Learning rate	&0.001	\\
Learning rate decay policy&	\makecell[r]{ReduceLROnPlateau (Reduce Learning \\Rate on Plateau) scheduler}	\\
	Learning rate factor&	0.8\\
	Patience	&30\\
	Min learning rate	&1.00E-07\\		
 \midrule  
% Network structure	&TorchMD-NET	\\
% 	Head number	&8	\\
% 	Layer number&	6	\\
% 	RBF number	&32	\\
% 	Activation function 	&SiLU	\\
% 	Embedding dimension&	128	\\
% \midrule  
Force weight	&0.8		\\
Energy weight	&0.2		\\
Noisy Nodes denoise weight	&0.1		\\
std. of torsion angles of rotatable bonds & 	20	\\
std. of CGN  & 	0.005	\\
% Dihedral angle noise scale(type: Gaussian)&	20		\\
% Coordinate noise scale(type: Gaussian)&	0.005		\\     
   \bottomrule
    \end{tabular}
    % \end{sc}
    % \end{footnotesize}
    % \end{center}
    % \vskip -0.1in
\end{table}
Hyperparameters for fine-tuning on MD17 are listed in \ref{table:app setting md17}. We test our model in two ways of data splitting. Correspondingly, there are two batch sizes proportional to the training data size. 

Hyperparameters for fine-tuning on MD22 and ISO17 are listed in \ref{table:app setting md22} and \ref{table:app setting iso17}, respectively.
For MD22 and ISO17, we apply a customized learning rate scheduler with a linear warmup and patience steps in the first training epoch.
We also provide different batch sizes and learning rates for the 7 tasks in MD22. 
\begin{table}[h!]
\centering
    \caption{\textbf{\thetable:} Hyperparameters for fine-tuning on MD22. }
    \label{table:app setting md22}
    % \vskip 0.15in
    % \begin{center}
    % \begin{footnotesize}
    % \begin{sc}
    \begin{tabular}{lr}
    \toprule
    Parameter & Value or description\\
     \midrule  
  Train/Val/Test Splitting & 80\%/10\%/10\% split same with \cite{coordnoneq2023} 	\\
  Batch size &	\makecell[r]{16 for double-walled-nanotube\\32 for the other 6 tasks}	\\
  \midrule  
Optimizer&	AdamW	\\
% Weight Decay&	0	\\
Epochs &100 \\
Max Learning rate	&\makecell[r]{0.0002 for AT-AT\\0.001 for Ac-Ala3-NHMe, DHA, AT-AT-CG-CG, Staychose\\ 0.005 for buckyball, double-walled-nanotube} \\
Learning rate decay policy&	Customized Cosine LRScheduler	 \\
Warm up steps	& 30\% steps of the first training epoch	\\
Patience steps	& 70\% steps of the first training epoch	\\
	Min learning rate	&1.00E-07\\		
 \midrule  
% Network structure	&TorchMD-NET	\\
% 	Head number	&8	\\
% 	Layer number&	6	\\
% 	RBF number	&32	\\
% 	Activation function 	&SiLU	\\
% 	Embedding dimension&	128	\\
% \midrule  
Force weight	&0.8		\\
Energy weight	&0.2		\\
   \bottomrule
    \end{tabular}
    % \end{sc}
    % \end{footnotesize}
    % \end{center}
    % \vskip -0.1in
\end{table}

\begin{table}[h!]
\centering
    \caption{\textbf{\thetable:} Hyperparameters for fine-tuning on ISO17.}
    \label{table:app setting iso17}
    % \vskip 0.15in
    % \begin{center}
    % \begin{footnotesize}
    % \begin{sc}
    \begin{tabular}{lr}
    \toprule
    Parameter & Value or description\\
     \midrule  
  Train/Val/Test Splitting &	Provided in http://quantum-machine.org/datasets/iso17.tar.gz \\
  Batch size &	256	\\
  \midrule  
Optimizer&	AdamW	\\
% Weight Decay&	0	\\
Epochs & 50 \\
Max Learning rate	&0.001	\\
Learning rate decay policy&	Customized Cosine LRScheduler	\\
Warm up steps	& 30\% steps of the first training epoch	\\
Patience steps	& 70\% steps of the first training epoch	\\
	Min learning rate	&1.00E-07\\		
 \midrule  
% Network structure	&TorchMD-NET	\\
% 	Head number	&8	\\
% 	Layer number&	6	\\
% 	RBF number	&32	\\
% 	Activation function 	&SiLU	\\
% 	Embedding dimension&	128	\\
% \midrule  
Force weight	&0.8		\\
Energy weight	&0.2		\\
   \bottomrule
    \end{tabular}
    % \end{sc}
    % \end{footnotesize}
    % \end{center}
    % \vskip -0.1in
\end{table}

\begin{table}[h!]
\centering
    \caption{\textbf{\thetable:} Hyperparameters for fine-tuning on QM9. }
    \label{table:app setting qm9}
    % \vskip 0.15in
    % \begin{center}
    % \begin{footnotesize}
    % \begin{sc}
    \begin{tabular}{lr}
    \toprule
    Parameter & Value or description\\
     \midrule  
Train/Val/Test Splitting	 & 110000/10000/remaining data	\\
Batch size & 	128	\\
  \midrule  
Optimizer	 & AdamW	\\
% Weight Decay & 	0	\\
Warm up steps & 	10000	\\
Max Learning rate	 & 0.0004	\\
Learning rate decay policy & 	Cosine	\\
	Learning rate factor	 & 0.8\\
	Cosine cycle length*	 & 300000 (500000) \\
  \midrule  		
% Network structure & 	TorchMD-NET	\\
% 	Head number	 & 8\\
% 	Layer number & 	8\\
% 	RBF number	 & 64\\
% 	Activation function  & 	SiLU\\
%  Embedding dimension	 & 256\\
%  Head&  \multirow{3}{*}{\makecell[c]{Applied according to\\ \href{https://github.com/torchmd/torchmd-net/issues/64}{https://github.com/torchmd/torchmd-net/issues/64}}}\\ 
% 	Standardize	&\\
% 	AtomRef&\\
% 	  \midrule  
Label weight	 & 1	\\
Noisy Nodes denoise weight	 & 0.1(0.2)	\\
% dihedral angle noise scale(type: Gaussian)	 & 2(Pretrain)	
std. of CGN  & 	0.005	\\
% Coordinate noise scale(type: Gaussian)	 & 0.005	\\   
   \bottomrule
    \end{tabular}
    % \end{sc}
    % \end{footnotesize}
    % \end{center}
    % \vskip -0.1in
\end{table}

\begin{table}[h!]
\centering
    \caption{\textbf{\thetable:} Hyperparameters for fine-tuning on LBA.}
    \label{table:app setting lba}
    % \vskip 0.15in
    % \begin{center}
    % \begin{footnotesize}
    % \begin{sc}
    \begin{tabular}{lr}
    \toprule
    Parameter & Value or description\\
     \midrule  
Train/Val/Test Splitting	 & 3507/466/490 (30 split); 3563/448/452(60 split)\\
Batch size & 	8	\\
  \midrule  
Optimizer	 & AdamW	\\
% Weight Decay & 	0	\\
Warm up steps & 	1000	\\
Max Learning rate	 & 0.0004	\\
Learning rate decay policy & 	Cosine	\\
	Learning rate factor	 & 0.8\\
	Cosine cycle length*	 & 40000 \\
  
   \bottomrule
    \end{tabular}
    % \end{sc}
    % \end{footnotesize}
    % \end{center}
    % \vskip -0.1in
\end{table}

Hyperparameters for fine-tuning on QM9 are listed in \ref{table:app setting qm9}. The cosine cycle length is set to be $500000$ for $\alpha$, $ZPVE$, $U_0$, $U$, $H$, $G$ and $300000$ for other tasks for fully converge.
 Notice that because the performance of QM9 and MD17 is quite stable for random seed, we will not run cross-validation. This also follows the main literature \cite{schutt2018schnet,schutt2021equivariant,liu2022spherical,ShengchaoLiu2022MolecularGP}.

Hyperparameters for fine-tuning on LBA are listed in \ref{table:app setting lba}. 

\added{In the ablation study on robustness in section 2.3, we use the RDKit toolkit to produce conformations of the molecules with 
 the Distance Geometry method and use Merck molecular force field (MMFF) to optimize the conformations, which are less accurate compared to those generated by DFT methods. Here, RDKit is selected as a representative empirical method for conformer generation, and other empirical methods, such as Openbabel's confab~\cite{o2011open}, can also be used.}
 
In the force accuracy experiment in section 2.4.1, the force estimation method for denoising is calculated according to the conditional score functions in \ref{tab:ff}. Since we only include one equilibrium $n=1$, the mixture model degenerates into a single Gaussian. Thus the conditional score function equals the score function and is exactly the force under the Boltzmann distribution (up to a constant). Note that all the quantities in \ref{tab:ff} can be specified, except the matrix $C$ in $\Sigma_{\sigma}$ and $\Sigma_{\sigma,\tau}$. Given the equilibrium, $C$ is determined and can be computed by least squares estimation. We utilize the ratio of residual to the estimation target $C_{error}=\frac{||\Delta x-C\Delta\psi||}{||\Delta x||}$ to measure the error of the least squares estimation of $C$.

Other metrics to measure the prediction error or correlation between $n$ pairs of ground truth values $y$ and predicted values $x$ are clarified as follows. The root mean square error (RMSE): $\sqrt{\frac{1}{n}\sum_{i=1}^n|x_i-y_i|^2}$.  The mean absolute error (MAE): $\frac{1}{n}\sum_{i=1}^n|x_i-y_i|$.
Pearson correlation coefficient: $\rho(x,y)\triangleq\frac{\sum_{i=1}^n(x_i-\bar{x})(y_i-\bar{y})}{\sqrt{\sum_{i=1}^n(x_i-\bar{x})^2}\sqrt{\sum_{i=1}^n(y_i-\bar{y})^2}}$, where $\bar{x}=\frac{1}{n}\sum_{i=1}^n x_i$, $\bar{y}=\frac{1}{n}\sum_{i=1}^n y_i$.
Spearman's rank correlation coefficient: $\rho(R(x),R(y))$, where $R(x_i),R(y_i)$ are ranks of $x_i$ and $y_i$.   %nyy

\subsection{\added{Computational Efficiency}}\label{sec:Computational Efficiency}
 \added{The practical real-world applicability of Frad, including its hardware requirements for deployment and speed advantages compared to existing methods, is presented in \ref{apptable:Hardware}. Considering both hardware and pre-training time, our method is among the fastest pre-training models available. This speed advantage is primarily due to our network structure, which is more lightweight and has fewer parameters compared to other Transformer models. Additionally, our efficiency benefits from a simpler input requirement, needing only 3D input modalities, whereas other methods may need to process both 2D and 3D inputs. Overall, Frad's efficient pre-training and superior performance make it a practical and scalable option for real-world applications.}
\begin{table}[h!]
\centering
\caption{\added{\textbf{\thetable}: Hardware Configuration and Pre-training Time for Different Pre-training Methods}}
\label{apptable:Hardware}
\begin{tabular}{lll}
\toprule
\textbf{Method}       & \textbf{Hardware Configuration} & \textbf{Pre-training Time} \\ \midrule
Frad                  & 1 NVIDIA A100 40GB              & 14h 56m 44s                \\ 
Coord                 & 3 2080Ti GPUs                   & 25 h                       \\ 
3D-EMGP               & 4 NVIDIA V100 GPUs              & 12 h                       \\ 
SE(3)-DDM             & 20 NVIDIA V100 GPUs             & 3-24 h                     \\ 
Transformer-M         & 4 NVIDIA A100 GPUs              & 75h                        \\ 
MoleBLEND \cite{yu2023multimodalblending}          & 4 NVIDIA A100 40GB              & 24 h                       \\ 
Uni-MOL \cite{zhou2023unimol}            & 8 NVIDIA V100 GPUs              & 20 h                       \\ 
MOL-AE \cite{yang2024molae}             & 1 NVIDIA A100 GPU               & 48 h                       \\ 
Grover \cite{rong2020self} (base)      & 250 NVIDIA V100 GPUs            & 60 h                       \\ 
Grover (large)        & 250 NVIDIA V100 GPUs            & 96 h                       \\ \bottomrule
\end{tabular}
\end{table}

\section{More related work}\label{app:Related work}
\subsection{Molecular pre-training methods}
Molecular pre-training has become a prevalent method for obtaining molecular representations due to the lack of downstream data. In general, molecular pre-training can be categorized by input data modality. 
Traditionally, emphasis has been placed on two primary modalities of input molecular data: 1D SMILES strings \cite{wang2019smiles,kim2021merged,zhang2021mg,xue2021x,guo2022multilingual,zhang2021mgbert} and 2D molecule graphs \cite{rong2020self,zhang2021motif,li2021effective,zhu2023dual,wang2022MolCLR,fang2022molecular,lin2022pangu,xia2023Molebert}. Most of the methods draw inspiration from the pre-training methods in CV and NLP, and implement masking and contrastive learning to learn molecular representations for 1D and 2D inputs.

Recently, a shift towards integrating 3D atomic coordinate positions as inputs has emerged, offering a more informative and inherent representation of molecules. Earlier methods treat 3D information as a complement to 2D input and learn to align the two modalities in a contrastive or generative way \cite{liu2021pre,li2022geomgcl,zhu2022unified,stark20223d,yu2023moleblend}. Recent methods have focused on devising self-supervised learning tasks tailored specifically for 3D geometry data, enabling direct representation learning on 3D inputs.
% \cite{fang2022geometry,zhou2023unimol,luo2022one,SheheryarZaidi2022PretrainingVD,ShengchaoLiu2022MolecularGP,jiao2022energy}. 

These 3D pre-training tasks can be divided into three categories: denoising, geometry masking, and geometry prediction.
Geometry masking methods design nontrivial masking tasks based on 3D molecules, such as \citet{fang2022geometry} propose to mask and predict the bond lengths and bond angles in molecular structures, and \citet{zhou2023unimol} mask and predict atom types based on noisy geometry. %ChemRL-GEM,unimol 
In predictive methods, targets closely related to conformation are predicted, such as \citet{luo2022one} aim to predict energy, \citet{fang2022geometry} predict atomic pairwise distances given bond lengths and bond angles, and \citet{wang2023automated} propose to predict bond lengths, bond angles and dihedral angles. 
% Moreover, coordinate denoising enjoys a force learning interpretation and thus helps downstream tasks. Therefore, we focus on coordinate denoising method in our work.

As for the denoising task, it involves corrupting and reconstructing 3D coordinates of the molecule \citep{SheheryarZaidi2022PretrainingVD,luo2022one,zhou2023unimol}. When utilizing coordinate Gaussian noise, denoising enjoys a force learning interpretation. Based on coordinate denoising, some works introduce equivariance to the molecular energy function. For instance, \citet{jiao2022energy} incorporates a Riemann-Gaussian noise to ensure E(3)-invariance in the energy function. Similarly, to maintain SE(3)-invariance of energy, \citet{ShengchaoLiu2022MolecularGP} propose denoise on pairwise atomic distance. It is worth noting that previous denoising methods rely on Gaussian-based noise, describing an isotropic molecular distribution that deviates from the true molecular distribution, resulting in limited sampling coverage and inaccurate forces. Consequently, our work introduces a novel approach that integrates chemical priors to capture a more realistic molecular distribution, an aspect often overlooked by previous denoising methods.
 
\subsection{Denoising and score matching}
Using noise to improve the generalization ability of neural networks has a long history\cite{sietsma1991creating,ChristopherMBishop1995TrainingWN}. Denoising autoencoders, as proposed by \citet{PascalVincent2008ExtractingAC,PascalVincent2010StackedDA}, introduce a denoising strategy to acquire robust and effective representations. In this context, denoising is interpreted as a method for learning the data manifold. In the domain of Graph Neural Networks (GNNs), researchers have demonstrated the performance improvement achieved through training with noisy inputs \cite{hu2019strategies,kong2020flag,sato2021random,godwin2021simple}. Specifically, Noisy Nodes~\cite{godwin2021simple} incorporates denoising as an auxiliary loss to alleviate over-smoothing and enhance molecular property prediction. 

Score matching is an energy-based generative model to maximize the likelihood for unnormalized probability density models whose partition function is intractable. The connection between denoising and score matching is established when the noise is standard gaussian~\cite{PascalVincent2011ACB}. This is successfully applied in generative modeling \cite{YangSong2019GenerativeMB,song2020improved,hu2019strategies,EmielHoogeboom2023EquivariantDF} and in energy-based modeling of molecules for pre-training~\citep{SheheryarZaidi2022PretrainingVD}. While both generative models and forces learning interpretation for denoising draw upon the findings of \cite{PascalVincent2011ACB}, they differ in their assumptions and aims in practice.

\newpage
\end{appendix}
\end{document}